\documentclass[10pt,twocolumn,twoside]{IEEEtran}
\usepackage[utf8]{inputenc} 
\usepackage[T1]{fontenc}    
\usepackage{url}            
\usepackage{booktabs,tabularx}       
\usepackage{amsfonts}       
\usepackage{nicefrac}       
\usepackage{microtype}      
\usepackage{amsthm}
\usepackage{algorithmicx,algorithm}
\usepackage[noend]{algpseudocode}
\usepackage{times}
\usepackage{epsfig}
\usepackage{graphicx}
\usepackage{amsmath}
\usepackage{amssymb}
\usepackage{amstext}
\usepackage{array}
\usepackage{bm}
\usepackage{multirow}
\usepackage[marginal]{footmisc}
\usepackage{ragged2e}
\usepackage{makecell}
\usepackage{color}

\newtheorem{myDef}{Definition}

\usepackage[pdftex, pagebackref, colorlinks, citecolor=blue, linkcolor=blue]{hyperref}

\usepackage{xspace}
\makeatletter
\DeclareRobustCommand\onedot{\futurelet\@let@token\@onedot}
\def\@onedot{\ifx\@let@token.\else.\null\fi\xspace}
\def\eg{\emph{e.g}\onedot} 
\def\ie{\emph{i.e}\onedot} 
 
\def\etc{\emph{etc}\onedot} 
 
\def\etal{\emph{et al}\onedot}
\makeatother

\begin{document}
	\title{
		Simultaneous Subspace Clustering and Cluster Number Estimating based on Triplet Relationship
	}
	
	\author{Jie Liang, Jufeng Yang, Ming-Ming Cheng, Paul L. Rosin, Liang Wang
		\thanks{J. Liang, J. Yang, and M.-M. Cheng are with College of Computer Science, Nankai University, Tianjin 300350, China (E-mail: liang27jie@163.com; yangjufeng@nankai.edu.cn; cmm@nankai.edu.cn).}
		\thanks{P. L. Rosin is with School of Computer Science and Informatics, Cardiff University, Wales, UK. (E-mail: Paul.Rosin@cs.cf.ac.uk).}
		\thanks{L. Wang is with the National Laboratory of Pattern Recognition, CAS
			Center for Excellence in Brain Science and Intelligence Technology, Institute
			of Automation, Chinese Academy of Sciences, Beijing 100190, China (E-mail:
			wangliang@nlpr.ia.ac.cn).
		}
		\thanks{A preliminary version of this work appeared at AAAI~\cite{yang2018AAAI}.}
	}
	
	\markboth{IEEE Transactions on image processing}
	{Liang \MakeLowercase{\etal}: Regular Paper}
	
	\maketitle
	
	
	\begin{abstract}
		
		In this paper we propose a unified framework to simultaneously discover the number of clusters and group the data points into them using subspace clustering.
		Real data distributed in a high-dimensional space can be disentangled into a union of low-dimensional subspaces, which can benefit various applications.
		To explore such intrinsic structure, state-of-the-art
		subspace clustering approaches often optimize a self-representation problem among all samples, to construct a pairwise affinity graph for spectral clustering.
		However, a graph with pairwise similarities lacks
		robustness for segmentation, especially for samples which lie on the intersection of two subspaces.
		To address this problem, we design a hyper-correlation based data structure termed as the \textit{triplet relationship}, which reveals high relevance and local compactness among three samples.
		The triplet relationship can be derived from the self-representation matrix, and be utilized to iteratively assign the data points to clusters.
		Three samples in each triplet are encouraged to be highly correlated and are considered as a meta-element during clustering, which show more robustness 
		than pairwise relationships
		when segmenting two densely distributed subspaces.
		Based on the triplet relationship, we propose a unified optimizing scheme to automatically calculate clustering assignments.
		Specifically, we optimize a model selection reward and a fusion reward by simultaneously maximizing the similarity of triplets from different clusters while minimizing the correlation of triplets from same cluster.
		The proposed algorithm also automatically reveals the number of clusters and fuses groups to avoid over-segmentation.
		Extensive experimental results on both synthetic and real-world datasets validate the effectiveness and robustness of the proposed method.
	\end{abstract}
	
	\begin{IEEEkeywords}
		Subspace clustering, triplet relationship, estimating the number of clusters, hyper-graph clustering
	\end{IEEEkeywords}
	
	\section{Introduction}
	
	\IEEEPARstart{W}{ith} the ability to disentangle latent structure of data in an unsupervised manner~\cite{wang2017exclusivity,peng2017constructing,Vidal2010A}, subspace clustering is regarded as an important technique in the data mining community and for various computer vision applications~\cite{jia2017subspace,xu2014clustering,zhu2017subspace,cao2015constrained}.
	Traditional subspace clustering methods approximate a set of high-dimensional data samples into a union of lower-dimensional linear subspaces~\cite{elhamifar2009sparse}, where each subspace usually contains a subset of the samples.
	
	In recent years, spectral clustering based methods have achieved state-of-the-art performance, taking a two-step framework as follows.
	First, by optimizing a self-representation problem~\cite{Vidal2010A,yangl0}, a similarity matrix (and also a similarity graph) is constructed to depict the relationship (or connection) among samples.
	Second, spectral clustering~\cite{shi2000normalized} is employed for calculating the final assignment based on eigen-decomposition of the affinity graph.
	Note that in practice, both the number of subspaces and their dimensionalities are always unknown~\cite{Vidal2010A,Wang2015DP}.
	Hence, the goals of subspace clustering include finding the appropriate number of clusters and grouping data points to them~\cite{Li2016Simultaneous,javed2017background}.

	Nevertheless, it is challenging to estimate the number of clusters in a unified optimization framework, since the definition of clusters is subjective, especially in the high-dimensional ambient space~\cite{Li2016Simultaneous}.
	Also, for samples which are of different clusters but near the intersection of two subspaces, they may be even closer than samples from same cluster.
	This may lead to a wrong estimation with several redundant clusters, namely over-segmentation problem.
	Therefore, most of the spectral based subspace clustering algorithms depend on a manually given and fixed number of clusters, which cannot be generalized for multiple applications~\cite{Vidal2010A}.
	
	Most clustering schemes group similar patterns into the same cluster by jointly minimizing the inter-cluster similarity and the intra-cluster dissimilarity~\cite{kumar2016hybrid}.
	Considering the complexity of the high-dimensional ambient data space, an effective way for estimating the number of clusters is to first map the raw samples into an intrinsic correlation space, namely similarity matrix, followed by an iterative optimization according to the local and global similarity relationships derived from the projection.
	Elhamifar~\etal~\cite{elhamifar2013sparse} propose that the permuted similarity matrix can be block-diagonal, where the number of blocks is identical to the number of clusters.
	Moreover, Peng~\etal~\cite{peng2017constructing} verify the intra-subspace projection dominance (IPD) of such a similarity matrix, which can be applied to self-representation optimizations with various kinds of regularizations.
	The IPD theory says that for two arbitrary samples from the same subspace and one from another, the generated similarities between the former samples are always larger than between latter ones in a noise-free system.
	
	Accordingly, considering an affinity graph derived from the similarity matrix~\cite{Nasihatkon2011Graph, zhan2018graph}, where the vertices denote data samples and the edge weights denote similarities, an automatic sub-graph segmentation can be greedily conducted via the following two steps inspired by the density based algorithms~\cite{Rodriguez2014Clustering}:
	1) constructing a proper number of initialized cluster centers by minimizing the weighted sum of all inter-cluster connections and maximizing the intra-cluster ones;
	2) merging the remaining samples to an existing cluster by maximizing the weighted connections between the sample and the cluster.
	
	Yet, there are also difficulties for these greedy iterative schemes.
	Since the ambient space can also be very dense~\cite{Scalable_Sparse}, any two points which are close by evaluating the pairwise distance may not belong to the same subspace, especially for samples near the intersection of two subspaces.
	Consequently, a hypergraph where each edge can be connected to more than two samples~\cite{Gao2013Laplacian,Kim2014Image} is proposed to solve the problem in traditional pairwise graphs.
	In this paper, we further introduce a novel data structure termed as the \textit{triplet relationship}, to explore the local geometry structure in projected space with hyper-correlations.
	Each triplet consists of three points and their correlations, which are considered as a meta-element for clustering.
	We require that all correlations are large enough, which indicates that the three points are strongly connected according to the IPD property~\cite{peng2017constructing}.
	
	In contrast to evaluating similarities using pairwise distances, the proposed triplet relationship demonstrates favorable performance due to the following two reasons.
	On one hand, it is more robust when partitioning the samples near the intersection of two subspaces since the mutual relevance among multiple samples can provide complementary information when calculating the local segmentation.
	On the other hand, the triplet evokes a hyper-similarity by efficiently counting the frequency of intra-triplet samples, which enables a greedy way to calculate the assignments.
	
	Based on the newly defined triplet, in this paper, we further propose a unified framework termed as the \textit{autoSC} to jointly estimate the number of clusters and group the samples by exploring the local density derived from triplet relationships.
	Specifically, we first calculate the self-representation for each sample via an off-the-shelf optimization scheme, followed by extracting the triplet relationship for all samples.
	Then, we greedily initialize a proper number of clusters via optimizing a new model selection reward, which is achieved by maximizing inter-cluster dissimilarity among triplets.
	Finally, we merge each of the remaining samples into an existing cluster to maximize the intra-cluster similarity by optimizing a new fusion reward.
	We also fuse groups to avoid over-segmentation.

	The main contributions of this paper are summarized as follows:
	\begin{itemize}
		\item{
			First, we define a hyper-dimensional triplet relationship which ensures a high relevance and density among three samples to reflect their local similarity. 
			We also validate the effectiveness of triplets and distinguish them against the standard pairwise relation.}
		\item{
			Second, we design a unified framework, \ie, autoSC, based on the intrinsic geometrical structures depicted by our triplet relationships. 
			The proposed autoSC can be used for simultaneous estimating the number of clusters and subspace clustering in a greedy way.}
	\end{itemize}

	Extensive experiments on benchmark datasets indicate that our autoSC outperforms the state-of-the-art methods in both effectiveness and efficiency.
	
	This paper is an extended version of our earlier conference paper~\cite{yang2018AAAI}, to which we enrich the contributions in the following five aspects:
	(1) We add detailed analysis of the proposed algorithm to distinguish it from comparative methods, for example, we add analysis and experimental validation on the computational complexity.
	%
	%
	(2) We provide a visualized illustration of the proposed autoSC for clearer presentation.
	(3) We propose a relaxation termed as the neighboring based autoSC (autoSC-N), which directly calculates the neighborhood relationship from raw data space and is more efficient than autoSC.
	(4) We conduct experiments on evaluating the influence of the parameter $m$ (number of preserved neighbors for each sample). 
	(5) We experimentally evaluate our method on 
	real-world application, \ie, motion segmentation, which also demonstrates the benefits of the proposed method.

	\section{Related Work}
	
	Automatically approximating samples in high-dimensional ambient space by a union of low-dimensional linear subspaces is considered to be a crucial task in computer vision~\cite{elhamifar2009sparse,schroff2015facenet,wang2014constraint,li2017structured,zhang2017flexible}.
	In this section, we review the related contributions in the following three aspects, \ie, self-representation calculation, estimating the number of clusters and hyper-graph clustering.

	\subsection{Calculating Self-Representation}
	
	To separate a collection of data samples which are drawn from a high-dimensional space according to the latent low-dimensional structure, 
	traditional self-expressiveness based subspace clustering method calculates a linear representation for each sample using the remaining samples as 
	a basis set or
	a dictionary~\cite{you2016scalable,Cheng_2016_CVPR}.
	Subspace clustering assumes that the set of data samples are drawn from a union of multiple subspaces, which can best fit the ambient space~\cite{elhamifar2013sparse}.
	There are numerous real applications
	satisfying this assumption with varying degrees of exactness~\cite{li2017simultaneous}, \eg, face recognition, motion segmentation, \etc.

	By solving an optimization problem with self-representation loss and regularizations, subspace clustering~\cite{wu2016ordered, Li2015Structured} calculates a similarity matrix where each entry indicates the relevance between two samples.
	Different regularizing schemes with various norms of the similarity matrix, \eg, $\ell_1$~\cite{elhamifar2009sparse}, $\ell_2$~\cite{Hu2014Smooth}, elastic net~\cite{You2016Oracle} or nuclear norm~\cite{fang2016robust}, can explore different intrinsic properties of the neighborhood space.
	There are mainly three types of the regularization terms, including sparse-oriented, densely-connected and mixed norms.

	Algorithms based on sparse-type norms~\cite{yangl0,rahmani2017innovation}, \eg, $ \ell_0 $ and $ \ell_1 $ norms, eliminate most of the non-zero values in the similarity matrix to ensure that there are no connections between samples from different clusters.
	Elhamifar and Vidal~\cite{elhamifar2009sparse} propose the sparse representation based on $ \ell_1$ norm optimization.
	The obtained similarity matrix recovers a sparse subspace representation but may not satisfy the graph connectivity if the dimension of the subspace is greater than three~\cite{Nasihatkon2011Graph}.
	In addition, the $ \ell_0 $ based subspace clustering methods aim to compute
	a sparse and subspace-preserving representation for each data sample. 
	Yang~\etal~\cite{yangl0} present a sparse clustering method with a regularizer based on the  $\ell_0$ norm  
	by using the proximal gradient descent method.
	Numerous alternative methods have been proposed for  
	$\ell_0 $ minimization while avoiding non-convex problems, \eg,
	orthogonal matching pursuit~\cite{Dyer2013Greedy} and nearest subspace neighbor~\cite{park2014greedy}.
	The scalable sparse subspace clustering by orthogonal matching pursuit (SSC-OMP) 
	method~\cite{you2016scalable} compares elements in each column of the dot product matrix to determine which positions of the similarity matrix should be non-zero.
	However, this general pairwise relationship does not reflect the sample correlation well,
	especially for data pairs in the intersection of two subspaces~\cite{purkait2017clustering}.

	In contrast, dense connection based methods, 
	such as smooth representation~\cite{Hu2014Smooth} with $ \ell_2 $ norm and low rank representation with nuclear norm based methods~\cite{Liu2010Robust,li2016structured}, 
	propose to preserve many non-zero values in the similarity matrix to ensure the connectivity among intra-cluster samples~\cite{guo2014spatial, Feng2014Robust,wang2016product}.
	For these densely connected frameworks~\cite{liu2016decentralized,Xiao2015FaLRR}, the similarity matrix is interpreted as a projected representation of raw samples.
	Each column of the matrix is considered as the self-representation of a sample, 
	and 
	should be dense for mapping invariance (also termed as the grouping effect~\cite{Lu2012Robust, Hu2014Smooth}).
	Low-rank clustering methods~\cite{liu2016deterministic,xu2017unified} solve a nuclear norm based optimization problem with the aim of generating a block diagonal solution with dense connections.
	However, the nuclear norm does not enforce subset selection well when noise exists, 
	and the self-representation is too dense to be an efficient feature.

	Neither a sparse nor dense similarity matrix reveals a comprehensive correlation structure among samples due to their conflicted nature~\cite{Wang2013Provable,Lai2014Efficient,Kim2016Robust}.
	Consequently, to achieve trade-off between sparsity and the grouping effect, numerous mixed norms, \eg, trace Lasso~\cite{Lu2015Correlation} and elastic net~\cite{You2016Oracle}, have been integrated into the optimization function.
	Nevertheless, the structure of the data correlations depends on the data matrix,
	and the mixed norm is not effective for structure selection.
	Therefore, this method does not perform consistently well on different applications.

	Recently, many frameworks that incorporate various constraints into the optimization function have been proposed to detect different intrinsic properties of the subspace~\cite{fang2016robust,Xu2015Reweighted,abin2018querying}.
	For instance, to handle sequential samples, Guo~\etal~\cite{Guo2013Spatial} explore the neighboring relationship by incorporating a new penalty, \ie, a lower triangular matrix with $ -1 $ on the diagonal and $ 1 $ on the second diagonal, to force consecutive columns in the similarity matrix to be closer.
	In this paper, based on the intrinsic neighboring relevance and geometrical structures depicted in the similarity matrix, we calculate triplet relationships to form a hyper-correlation constraint of the clustering system.
	We validate the robustness of the proposed triplet relationship on top of different similarity matrices with various intrinsic properties.

	\subsection{Estimating the Number of Clusters}

	Most of the real applications in computer vision require estimating the number of clusters, according to the latent distribution of data samples~\cite{elhamifar2009sparse}.
	To solve this problem, three main techniques exists: singular-based Laplacian matrix decomposition, density-based greedy assignment and hyper-graph based segmentation.

	Singular-based Laplacian matrix decomposition is common in subspace clustering~\cite{wang2017constrained} and spectral clustering~\cite{yang2015multitask} due to the availability of the similarity matrix.
	Liu \etal~\cite{Liu2013Robust} propose a heuristic estimator inspired by the block-diagonal structure of the similarity matrix~\cite{Feng2014Robust}.
	Specifically, they estimate the number of clusters by counting the small singular values of a normalized Laplacian matrix which should be smaller than a given cut-off threshold.
	These singular based methods~\cite{Favaro2011A,elhamifar2009sparse} are dependent on a large 
	gap between singular values, which is limited to applications in which the subspaces are sparsely distributed in the ambient space.
	%
	%
	Meanwhile, the matrix decomposition process is time-consuming when extended to large scale problems.
	Recently, Li~\etal propose SCAMS~\cite{Li2014SCAMS,li2017simultaneous} which estimates the number of clusters by minimizing the rank of a binary relationship matrix encoding the pairwise relevance among all data samples.
	Simultaneously, they incorporate a penalty term on the clustering cost by minimizing the Frobenius inner product of the similarity matrix and binary relationship matrix.
	
	Density based methods~\cite{Ester1996A} greedily discover both the optimal number of clusters and the assignments of data to the clusters according to the local and global densities which are calculated by the pairwise distances in ambient space.
	Rodriguez \etal \cite{Rodriguez2014Clustering} automatically cluster samples based on the assumption that each cluster center is characterized by a higher density in the weight space than all its neighbors, while different centers should be far apart enough to avoid redundancy.
	Specifically, for each sample, its Euclidean-based local density and the distance to any points with higher densities are iteratively calculated and updated.
	In each iteration, the algorithm finds a trade-off between the density of cluster centers and the inter-cluster distance to update the assignments.
	Wang \etal~\cite{Wang2015DP} employ the Bayesian nonparametric method based on a Dirichlet process, and propose DP-space, which exploits a trade-off between data fitness and model complexity.
	DP-space is more tolerate to noisy and outlier values than the alternative algebraic and geometric solutions.
	Recently, correlation clustering (CC)~\cite{Beier2015Fusion} first constructs an undirected graph with positive and negative edge weights, followed by minimizing the sum of cut weights during the segmenting process.
	Sequentially, the clustering assignments can be optimized with a greedy scheme.
	Nevertheless, most of these density based algorithms are limited to pairwise correlation when evaluating the similarity of data samples, which is not robust for densely distributed subspaces.
	
	\begin{figure*}[t]
		\centering
		\includegraphics[width=0.98\textwidth]{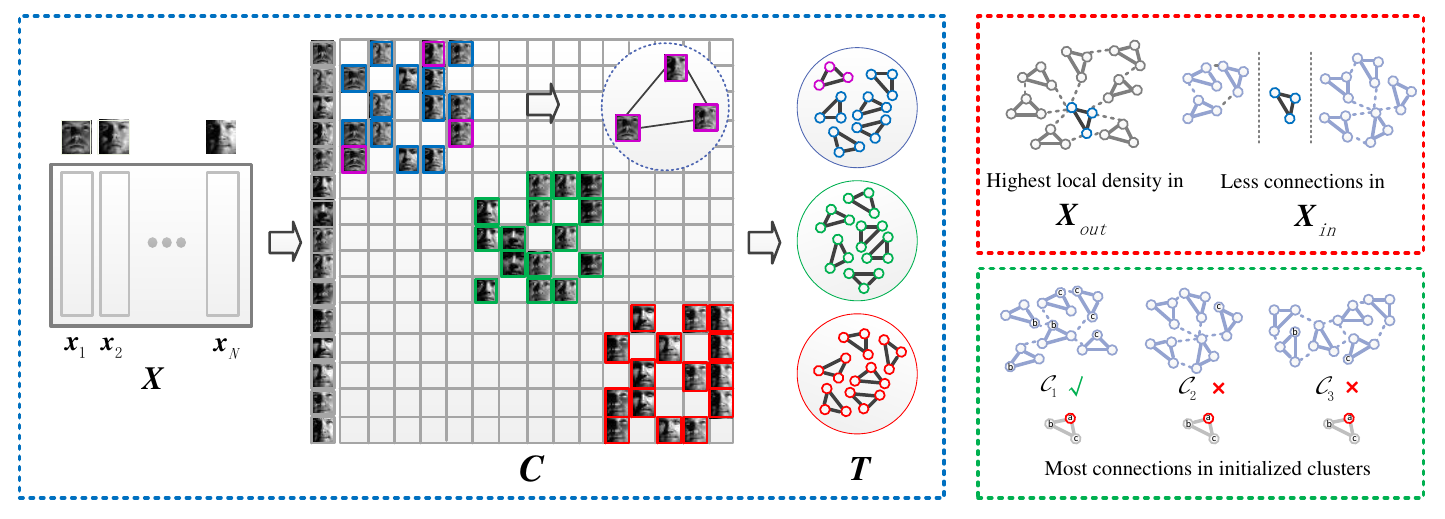}
		\vspace{-1em}
		\caption{Overview of the proposed autoSC. 
			The algorithm is composed of three steps, \ie, calculating triplet relationships $\bm{T}$ (blue dashed box), estimating the number of clusters via model selection reward (black) and finishing the clustering assignment via fusion reward (green).
			Given similarity matrix $\bm{C}$ derived from self-representation schemes, we illustrate an example of the triplet which is composed of the three samples shown with magenta frames.
			These samples induce a high local correlation and should be grouped into the same cluster.
			By optimizing the fusion reward, sample `a' is assigned into $ \mathcal{C}_1 $ since $ \mathcal{C}_1 $ has most connections to the triplet which involves `a'.
			\label{pipeFigure}
		}
	\end{figure*}

	\subsection{Hyper-graph Clustering}
	
	To tackle the limitations of the pairwise relation based methods, the hyper-graph relation~\cite{lutensor,Purkait2014Clustering,li2017graph} 
	is proposed and the related literature follows two different directions.
	Some transform the hyper-correlation into a simpler pairwise graph~\cite{Gao2013Laplacian,Sch2006Learning}, followed by a standard graph clustering method, \eg, normalized cut~\cite{shi2000normalized}, to calculate the assignments.
	Besides, other methods~\cite{Liu2010Robust,Li2016Simultaneous} explore a generalized way of extending the pairwise graph to the hyper-graph or hyper-dimensional tensor analysis.
	For instance, Li~\etal propose a tensor affinity variant of SCAMS, \ie, SCAMSTA~\cite{Li2016Simultaneous}, which exploits the higher order mathematical structures by providing multiple groups of nodes in the binary matrix derived from an outer product operation on multiple indicator vectors.
	However, estimating the number of clusters from the rank of the affinity matrix only works well for the ideal case, and can hardly be extended to complex applications since the noise can have a significant impact on the rank of affinity matrix.
	
	In this paper, we estimate the number of clusters by initializing the cluster centers with maximum inter-cluster dissimilarities and also maximum local densities.
	We calculate the initialization according to the local correlations reflected by the proposed triplet relationships, where each of them depicts a hyper-dimensional similarity among three samples and easily-evaluated relevances to other triplets.
	Both theoretical analysis on triplets as well as the experimental results demonstrate the effectiveness of the proposed method.

	\section{Methodology}
	
	\subsection{Preliminary}
	
	The main notations in the manuscript and the corresponding descriptions are shown in Table~\ref{notation}.
	Given a set of $ N $ data samples $ \bm{X} = \{\bm{x}_i\in \mathbb{R}^D\}_{i=1}^N $ lying in $K$ subspaces $\{S_i\}_{i=1}^K$ where $D$ denotes the dimensionality of each sample, 
	spectral based subspace clustering usually takes a two-step approach to calculate the clustering assignment.
	First, it learns a self-representation for each sample to disentangle the subspace structure.
	The algorithm then employs spectral clustering~\cite{shi2000normalized} on the learned similarity graph derived from $ \bm{C} $ for final assignments.
	Note that in practice, both the number of subspaces $ K $ and their dimensions $ \{d_j\}_{j=1}^K $ are always unknown~\cite{Vidal2010A,Wang2015DP}.
	Hence, the goals of subspace clustering include finding the appropriate $ K $ and assigning data points into $ K $ clusters~\cite{elhamifar2013sparse,Li2016Simultaneous}.

	In this paper, inspired by the block-diagonal structure of the similarity matrix~\cite{lee2015membership}, we propose to simultaneously estimate the number of clusters and assign samples into each cluster in a greedy manner.
	We design a novel meta-sample that we call a \textit{triplet relationship}, followed by optimizing both a model selection reward and a fusion reward for clustering.

	\subsection{Learning the Self-Representation}

	To explore the neighboring relationship in the ambient space $ \bm{X} = \{\bm{x}_i\in \mathbb{R}^D\}_{i=1}^N $, typical subspace clustering methods first optimize a linear representation of each data sample using the remaining dataset as a dictionary.
	Specifically, spectral based subspace clustering calculates a similarity matrix $ \bm{C}\in\mathbb{R}^{N\times N} $ by solving a self-representation optimization problem as follows:
	\begin{equation}\label{intro}
	\min_{\bm{C}}\   L(\bm{X}\bm{C},\bm{X})+\lambda\lVert{\bm{C} }\rVert_{\xi},
	\end{equation}
	where $ L(\cdot,\cdot): \mathbb{R}^{N\times D} - \mathbb{R}^{N\times D} \to\mathbb{R}^+$ denotes the reconstruction loss, $\lambda$ is the trade-off parameter and $ \lVert{\cdot }\rVert_{\xi} $ denotes the regularization term  where different $ \xi $'s lead to various norms~\cite{Vidal2010A,yangl0}, \eg,
	$\lVert{\bm{C} }\rVert_1$~\cite{elhamifar2009sparse,elhamifar2013sparse},
	$\lVert{\bm{C} }\rVert_*$~\cite{Liu2013Robust,Feng2014Robust},
	$\lVert{\bm{C} }\rVert_F^2$~\cite{Lu2012Robust}, 
	or many kinds of mixed norms like 
	trace Lasso~\cite{Lu2015Correlation} or 
	elastic net~\cite{You2016Oracle}.

	The $ \bm{C} $ in~\eqref{intro} can be explained as a new representation of $ \bm{X} $, where each sample $ \bm{x}_i\in\mathbb{R}^D $ is mapped to $ \bm{c}_i\in\mathbb{R}^N $.
	Furthermore, $\bm{C}$ is a pairwise distance matrix where each entry $ c_{ij} $ reflects the similarity between two samples $ \bm{x}_i $ and $ \bm{x}_j $.
	Nevertheless, the pairwise distance reflects poor discriminative capacity on partitioning samples near the intersection of two subspaces.
	To handle this problem, in this paper, we explore a higher-dimensional similarity called Triplet relationship, which is based on a greedy combination of pairwise distances reflected by $\bm{C}$.

	\subsection{Discovering Triplet Relationships}

	\begin{table}[t]
		\centering
		\caption{Summary of the main notations in manuscript and the corresponding descriptions.}
		\label{notation}
		\begin{tabular}{p{1.5cm}<{\centering}p{6.2cm}}
			\toprule
			\textbf{Notation}&  \textbf{Description}\\
			\midrule
			$ \bm{x},\bm{X} $& data sample, a set of samples\\
			$ \bm{X}_{-j} $& samples in $\bm{X}$ without $\bm{x}_j$\\
			$S$& a subspace in the ambient space\\
			$ \bm{\tau},\bm{T} $& triplet, a set of triplets\\
			$ d, D $& dimensionality of subspace, dimensionality of sample\\
			$ n, N $& number of triplets, number of samples \\
			$ K, \widetilde{K}, \widehat{K} $& real number of clusters, initialized number of cluster centers, estimated number of clusters \\
			$\bm{C}$& similarity matrix derived from subspace representation method \\
			$ \bm{C}^* $& binary similarity matrix preserving the top $ m $ values in each row of $\bm{C}$ and modifying them to $ 1 $\\
			$c_{ij}$&  $ij$-th entry of the similarity matrix $\bm{C}$\\
			$\bm{c}_i$&  $i$-th column of $\bm{C}$\\
			$ \bm{N}_m(\bm{x})$&  set of $ m $ nearest neighbors of $ \bm{x} $ \\
			$ \mathcal{C} $& cluster center which contains part of samples in a subspace \\
			$ \bm{T}^I_{in},\bm{T}^I_{out} $ & set of triplets which are already/not assigned into clusters in the $ I $-th iteration\\
			$ \bm{X}^I_{in},\bm{X}^I_{out} $& set of samples which are already/not assigned into clusters in the $ I $-th iteration, preserving the frequency\\
			$ R_m(\mathcal{C}) $&  model selection reward of $ \mathcal{C} $ \\
			$ R_f^i(\mathcal{C}_i|\bm{x}) $&  fusion reward that $ \bm{x} $ being fused into $ \mathcal{C}_i $ \\
			$ s_{ij} $&  connection score of $ \bm{x}_{i} $ toward $ \bm{x}_j $ \\
			$ \rho(\bm{\tau},\bm{X}) $&  local density of $ \bm{\tau} $ against $ \bm{X} $\\
			$ \mathcal{G}_i,\bm{\mathcal{G}} $ &one of the result groups, set of the result groups\\ 
			NC$ _e $&  deviation rate between the estimated $ \widehat{K} $ and $ K $ \\
			$ \mathcal{A} $& error rate of the triplets \\
			\bottomrule
		\end{tabular}
	\end{table}
	
	Given the similarity matrix $\bm{C}$ where each entry $ c_{ij} $ reflects the pairwise relationship between $ \bm{x}_i $ and $ \bm{x}_j $, we propose to find the neighboring structure in a greedy way.
	%
	%
	For each data sample $ \bm{x}_j\in S $, subspace clustering algorithms calculate a projected adjacent space based on the self-expressive property, \ie, each data sample can be reconstructed by a linear combination of other points in the dataset~\cite{elhamifar2013sparse,Belkin2001Laplacian}.
	Therefore, $ \bm{x}_j $ is represented as
	\begin{equation}
	\label{eachsample}
	\bm{x}_j = \bm{X}_{-j}\times[c_{1j},c_{2j},\cdots,c_{(j-1)j}, c_{(j+1)j}, \cdots, c_{Nj}]
	\end{equation}
	where $ \bm{X}_{-j} $ includes samples except for $\bm{x}_j$, which is considered as a self-expression dictionary for representation.
	In addition, $ [c_{1j},c_{2j},\cdots,c_{(j-1)j}, c_{(j+1)j}, \cdots, c_{Nj}] $ records the coefficients of such combination system.
	With the regularization from various well-designed norms on $ \bm{c}_j $, the optimized result of~\eqref{eachsample} is capable of preserving only 
	linear
	combinations of samples in $ S $ while 
	eliminating others.
	Inspired by \cite{peng2017constructing, park2014greedy}, for each sample $\bm{x}_j$, we first collect its $ m $ nearest neighbors, i.e. those with the top $ m $ coefficients in $ \bm{c}_j $.
	The $ m $ nearest neighbors are defined as follows.
	
	\begin{myDef}
		\label{mNN}
		{\bf ($ \bm{m}$ Nearest Neighbors) \ }  \textit{Let $ \bm{N}_m(\bm{x}_j) $ $\in$ $ \mathbb{R}^{1\times m}$ denote the $ m $ nearest neighbors for data point $ \bm{x}_j $. Let:
		\begin{equation}
		\bm{N}_m(\bm{x}_j) = \arg\max_{\{\bm{x}_{\bm{i}_l}\}} \sum_{l=1}^{m} |c_{\bm{i}_l,j}|.
		\label{Nm}
		\end{equation}}
		where $\bm{i}_l$ denotes the set of 
		indices for the nearest neighbors, and $ c_{\bm{i}_l,j} $ denotes the coefficient between $ \bm{x}_{\bm{i}_l} $ and $ \bm{x}_j $.
	\end{myDef}
	
	According to Definition~\ref{mNN}, we obtain $ \bm{N}_m(\bm{x}_j) $ for $\bm{x}_j$ which contains the $ m $ samples with the $ m $ largest coefficients in $ \bm{c}_j $.
	%
	%
	The number of preserved neighbors, \ie, the parameter $m$, reflects the intrinsic dimension of the low-dimensional subspaces~\cite{NIPS_Manifold,Li2009Learning}, which we empirically evaluated in the experiment section.
	Based on the $ m $ nearest neighbors, we define the triplet relationship to explore the local hyper-correlation among samples.
	\begin{myDef}
		\label{triplet_relationship}
		{\bf (Triplet Relationship) \ }  \textit{A triplet $ \bm{\tau} $ includes three samples, \ie, $ \bm{\tau} =\{\bm{x}_i,\bm{x}_j,\bm{x}_k\} $, and their relationships, if and only if $ \bm{x}_i,\bm{x}_j $ and $\bm{x}_k $ satisfy:
			\begin{equation}
			\bm{1}_{\bm{x}_i\in\bm{N}_m(\bm{x}_j)} \times \bm{1}_{\bm{x}_j\in\bm{N}_m(\bm{x}_k)} \times \bm{1}_{\bm{x}_k\in\bm{N}_m(\bm{x}_i)}=1,
			\label{triplet}
			\end{equation}
			where $\bm{1}_{\bm{x}\in\bm{N}_m}$ denotes the indicator function which equals $1$ if $ \bm{x}\in\bm{N}_m $ and $0$ otherwise.
		}
	\end{myDef}
	Based on Definition~\ref{triplet_relationship}, we obtain $ n $ triplets where we always have $ n>N $, \ie, each sample is included in multiple triplet relationships.
	For clarity of presentation, we define a triplet matrix $ \bm{T}\in\mathbb{R}^{n\times 3} $ for data samples $\bm{X}$, where each row of  $ \bm{T}$ records the indices of a samples in a triplet $ \bm{\tau} =\{\bm{x}_i,\bm{x}_j,\bm{x}_k\} $.

	Compared against the traditional pairwise relationship evoked from $ \bm{C} $, the triplet incorporates complementary using the constraint in~\eqref{triplet}, which shows more robust capacity in partitioning samples near the intersection of two subspaces.
	Each triplet depicts a local geometrical structure which enables a better performance to estimate the density of each sample.
	Furthermore, the overlapped samples in multiple triplets reflect a global hyper-similarity among each other, which can be measured efficiently.
	Therefore, based on the triplet relationship, we can jointly estimate the number of subspaces and calculate the clustering assignment in a greedy manner.

	\subsection{Modeling Clustering Rewards}

	Given $\bm{X}$, we iteratively group data samples into clusters, \ie, $ \{\mathcal{C}_i\}_{i=1}^{\widehat{K}} $, where $ \widehat{K} $ denotes the estimated number of subspaces.
	According to the greedy strategy, in the $ I $-th iteration, the triplet $ \bm{T} $ is divided into two subsets, \ie, ``in-cluster'' triplets $ \bm{T}_{in}^I $ which are already assigned into clusters, and ``out-of-cluster'' triplets $ \bm{T}_{out}^I $ which are still to be assigned in the subsequent iterations.
	For clearer presentation, we reshape both matrices $ \bm{T}_{in}^I\in\mathbb{R}^{p\times 3} $ and $ \bm{T}_{out}^I\in\mathbb{R}^{q\times 3} $ to vectors $ \bm{X}_{in}^I\in\mathbb{R}^{3p} $ and $ \bm{X}_{out}^I \in\mathbb{R}^{3q}$.
	In each iteration, we propose to optimize two new rewards, \ie, the model selection and the fusion reward, to simultaneously estimate the number of clusters and merge samples into respective cluster.

	\begin{myDef}
		{\bf (Model Selection Reward) \ }  \textit{Given $\bm{X}_{in}^I$ and $\bm{X}_{out}^I$ in the $I$-th iteration, the model selection reward $ R_m(\mathcal{C}) $ for each initialized cluster $ \mathcal{C} $ in $ \{\mathcal{C}_i\}_{i=1}^{\widehat{K}} $ is defined as:
			\begin{equation}
			R_m(\mathcal{C})=\sum_i \sigma(\mathcal{C}_i|\bm{X}_{out}^I)-\lambda_m\sum_i \sigma(\mathcal{C}_i|\bm{X}_{in}^I),
			\label{modelselection}
			\end{equation}
			where $ \sigma(\mathcal{C}|\bm{X}) $ is a counting function on the frequency that $\bm{x} \in \bm{X}$ for all $ \bm{x}\in\mathcal{C}_i $, $\lambda_m $ denotes the trade-off parameter.
		}
	\end{myDef}
	
	By maximizing the model selection reward $ R_m(\mathcal{C}) $, we generate the initialized cluster $ \{\mathcal{C}_i\}_{i=1}^{\widehat{K}} $ which has the following two advantages, where $\widehat{K}$ is the estimated number of clusters (see Fig.~\ref{pipeFigure} for visualization).
	Firstly, the local density of sample $ \bm{x}\in \mathcal{C} $ is high, \ie, $ \bm{x} $ has a large amount of correlated samples in $ \bm{X}_{out} $, which enables many to be merged in the next iteration.
	Secondly, each $ \mathcal{C} $ has little correlation with samples in $ \bm{X}_{in} $, which eliminates the overlap of any inter-clusters.
	Consequently, we can simultaneously estimate $ \widehat{K} $ and initialize the clusters by optimizing the model selection reward $ R_m $.
	
	\begin{myDef}
		{\bf (Fusion Reward) \ }  \textit{Given the initialized clusters $ \{\mathcal{C}_i\}_{i=1}^{\widehat{K}} $, the fusion reward is defined as the probability that $ \bm{x}_j\in\bm{X}_{out} $ is assigned into $ \mathcal{C}_i $:
			\begin{equation}
			\begin{split}
			R_f^i(\mathcal{C}_i|\bm{x}_j\in\bm{X}_{out}) =
			\sigma(\bm{x}_j|\mathcal{C}_i)+\lambda_f   \sigma(\bm{N}_m(\bm{x}_j)|\bm{N}_m(\mathcal{C}_i)),
			\label{fusion}
			\end{split}
			\end{equation}
			where $ \bm{N}_m(\bm{x}_j) $ denotes the $ m $ nearest neighbors of $ \bm{x}_j $ and $ \bm{N}_m(\mathcal{C}_i) $ denotes the set of $ m $ nearest neighbors of samples in $ \mathcal{C}_i $, $ \lambda_f $ denotes the trade-off parameters.
		}
	\end{myDef}
	
	In the optimization procedure, we calculate $ \widehat{K} $ fusion rewards $ \{R_f^i\}_{i=1}^{\widehat{K}} $ for each $ \bm{x}_j $, which represent the probabilities that $ \bm{x}_j $ is assigned into clusters $ \{\mathcal{C}_i\}_{i=1}^{\widehat{K}} $, respectively.
	We then merge $ \bm{x}_j $ into the cluster with the largest fusion reward, and move $ \bm{x}_j $ from $ \bm{X}_{out}  $ to $ \bm{X}_{in} $.

	\begin{algorithm}[t]
		\caption{\bf : Automatic Subspace Clustering (autoSC)\label{autoSC}}
		\label{algorithm}
		\hspace*{0.02in} {\bf Input:}
		$ \bm{X}=[\bm{x}_1,\cdots,\bm{x}_N]\in \mathbb{R}^{D\times N}$, $m$.
		\begin{algorithmic}[1]
			\State Normalize the magnitudes by $\bm{x}_i\leftarrow\bm{x}_i/\lVert \bm{x} \rVert_2$;
			\State Calculate the similarity matrix $ \bm{C} $ by~\eqref{intro};
			\For{$ i=1:N $}
			\State Calculate the $ m $ Nearest Neighbors $\bm{N}_{m} (\bm{x}_i)$ by~\eqref{Nm};
			\EndFor \State {\bf end for}
			\State Generate the triplet matrix $\bm{T}\in\mathbb{R}^{n\times 3}$ by~\eqref{tau_calculation};
			\State Reshape $ \bm{T} $ to $\bm{X}_{out}\in\mathbb{R}^{3n}$, $ \bm{X}_{in}=\varnothing $;
			\State $\widetilde{K}=1$;
			\State Calculate $ \bm{\tau}_{ini}^{\widetilde{K}} $ by~\eqref{tauini};     \While{$\rho(\bm{\tau}_{ini}^{\widetilde{K}},\bm{X}_{out})>\rho(\bm{\tau}_{ini}^{\widetilde{K}},\bm{X}_{in})$}
			\State $\mathcal{C}_{\widetilde{K}}=\bm{\tau}_{ini}^{\widetilde{K}}$;
			\State $\mathcal{C}_{\widetilde{K}}=\mathcal{C}_{\widetilde{K}}\cup\{\bm{\tau}^*\}$ where $ \bm{\tau}^* $ is calculated by~\eqref{taustar};
			\State $\bm{X}_{in}=\bm{X}_{in}\cup\bm{\tau}_{ini}^{\widetilde{K}}\cup\{\bm{\tau}^*\}$; 
			\State $\bm{X}_{out}=\bm{X}_{out}/(\bm{\tau}_{ini}^{\widetilde{K}}\cup\{\bm{\tau}^*\})$;
			\State $\widehat{K}=\widetilde{K}+1$;
			\State Calculate $ \bm{\tau}_{ini}^{\widetilde{K}} $ by~\eqref{tauini};
			\EndWhile \State {\bf end while}
			\State Merge $\mathcal{C}_i$ and $\mathcal{C}_j$ if we have~\eqref{sijcondition}; Get $ \widehat{K} $ clusters;
			\For{$ j=1:|\bm{X}_{out}| $}
			\State Calculate $ \mathcal{C}^* $ for $\bm{x}_j$ by~\eqref{rest};
			\EndFor \State {\bf end for}
		\end{algorithmic}
		\hspace*{0.02in} {\bf Output:}
		The cluster assignment $\{\mathcal{C}_i\}_{i=1}^{\widehat{K}}$.
	\end{algorithm}
	
	\subsection{Automatic Subspace Clustering Algorithm}
	
	The first triplet for initializing a new cluster is chosen to have maximal local density.
	%
	The local density $ \rho $ is defined as follows.
	
	\begin{myDef}
		\label{localdense}
		{\bf (Local Density) \ }  \textit{The local density $ \rho $ of the triplet $ \bm{\tau} $ regarding to the $ \bm{X}_{out} $ is defined as follows:
			\begin{equation}
			\rho ( \bm{\tau},\bm{X}_{out})= \sum_{j=1}^{|\bm{n}|}\sigma(\bm{x}_{\bm{n}_j}|\bm{X}_{out}),
			\end{equation}
			where $ \bm{x}_{\bm{n}_j} $ denotes the sample in the current triplet $ \bm{\tau} $ and $\bm{n}$ is the set of their indexes, $ |\bm{n}| $ denotes the scale of $\bm{n}$.
		}
	\end{myDef}
	
	Also, to measure the hyper-similarity between samples and determine the optimal triplet to merge into the initialized clusters, we define the connection score $ s $ as follows.
	
	\begin{myDef}
		{\bf (Connection Score) \ }  \textit{The connection score $ s $ between samples $ \bm{x}_i $ and $ \bm{x}_j $ is defined as:
			\begin{equation}
			s(\bm{x}_i,\bm{x}_j)=f\left(\bm{x}_i\bigg|\displaystyle\bigcap_{k=1}^{n'}\left(\bm{1}_{\bm{x}_j\in\bm{\tau}_k}\times \bm{\tau}_k\right)\right),
			\label{score}
			\end{equation}
			where $ \bm{1}_{\bm{x}_j\in\bm{\tau}_k} $ is equal to $ 1 $ when $ \bm{x}_j\in\bm{\tau}_k $ and $ 0 $ otherwise, $ n' $ is the number of all triplets in $ \bm{T}_{out} $.
		}
	\end{myDef}
	
	
	We greedily optimize the proposed model selection reward $  R_m  $ and fusion reward $ R_f $ in autoSC to simultaneously estimate the number of clusters and generate the segmentation among samples:
	\begin{equation}
	\begin{split}
	&\max_{\bm{\mathcal{G}},\widehat{K}} \
	\sum_{k=1}^{\widehat{K}}R_m(\mathcal{G}_k) + \lambda\sum_{k=1}^{\widehat{K}}R_f(\mathcal{G}_k|\bm{X}), \\
	&s.t.\ \mathcal{G}_k\bigcap\mathcal{G}_{k'\ne k}=\varnothing,
	\bigcup_{k=1}^{\widehat{K}}\mathcal{G}_k=[1,\cdots,N],
	\end{split}
	\label{aoverallmath}
	\end{equation}
	where $ \bm{\mathcal{G}}=\{\mathcal{G}_1,\cdots,\mathcal{G}_{\widehat{K}}\} $ denotes the set of the result groups, $ \widehat{K} $ is the estimated number of clusters and $ [1,\cdots,N] $ denotes the universal ordinal set of samples.
	
	We present the proposed autoSC in Fig.~\ref{pipeFigure} and Algorithm~\ref{algorithm}.
	Specifically, the optimization includes three steps:
	1) generating the triplet relationships $ \bm{T} $ from the similarity matrix $ \bm{C} $;
	2) estimating the number of clusters $ \widehat{K} $ and initializing the clusters $ \mathcal{C} $;
	3) assigning the samples $ \bm{x}\in\bm{X}_{out} $ into proper cluster.
	
	\vspace{1em}
	\noindent\textbf{Calculating Triplets:}
	The similarity matrix $ \bm{C} $ reflects the correlations among samples~\cite{elhamifar2009sparse}, where larger values demonstrate stronger belief for the correlation between samples. 
	For instance, $ c_{ij}>c_{ik} $ indicates a larger probability for $\bm{x}_i $ and $ \bm{x}_j $ being in the same cluster over $ \bm{x}_i $ and $ \bm{x}_k $.
	Accordingly, we explore the intrinsic local correlations among samples by the proposed triplets derived from $\bm{C}$.
	
	Many subspace representations guarantee the mapping invariance via a dense similarity matrix $\bm{C}$.
	However, the generation of triplets relies only on the strongest connections to avoid the wrong assignment.
	Therefore, for each column of $ \bm{C} $, \ie, $ \bm{c}_i $, we preserve only the top $ m $ values which are then modified to $ 1 $ for a new binary similarity matrix $ \bm{C}^* $.
	%
	
	Then, we extract each triplet from $ \bm{C}^* $ by the following function:
	\begin{equation}
	\begin{split}
	&\bm{\tau}=\{\bm{x}_{\bm{n}_1},\bm{x}_{\bm{n}_2},\bm{x}_{\bm{n}_3}\}\in\bm{T},\\ &\text{if\ and\ only\ if}:\ \  c_{\bm{n}_1\bm{n}_2}^*\times c_{\bm{n}_2\bm{n}_3}^*\times c_{\bm{n}_3\bm{n}_1}^*=1,
	\end{split}
	\label{tau_calculation}
	\end{equation}
	where $ c^*_{xy} $ denotes the $ xy $-th value of $ \bm{C}^* $.
	Note each sample $ \bm{x} $ can appear in many triplets.
	Therefore, we consider each $ \bm{\tau} $ as a meta-element in the clustering, which improves the robustness due to the complementarity constraints.
	
	\vspace{1em}
	\noindent\textbf{Initializing Clusters:}
	%
	In the $I$-th iteration, we first determine an initial triplet (termed as $ \bm{\tau}_{ini}^I $) from $ \bm{T}_{out} $ to initialize the cluster $ \mathcal{C} $,
	followed by merging the most correlated samples of $ \bm{\tau}_{ini}^I $ into each $ \mathcal{C} $.
	
	Following~\cite{Rodriguez2014Clustering}, we initialize a new cluster using $ \bm{\tau}_{ini}^I $ with highest local density:
	\begin{equation}
	\bm{\tau}_{ini}^I=\arg\max_{\bm{\tau}}\ \ \rho ( \bm{\tau},\bm{X}_{out}^I),
	\label{tauini}
	\end{equation}
	where $ \rho $ calculates the local density defined in Definition~\ref{localdense}.
	The high local density of the triplet reflects the most connections between $ \bm{\tau}_i $ and other triplets, which produces the most connections between $ \bm{x}_{\bm{n}_j^i} $ and other samples in $ \bm{X}_{out}^I $.
	
	Once the initialized triplet $ \bm{\tau}_{ini}^I $ is determined, we iteratively extend the initialized cluster $ \mathcal{C} $ by fusing the most confident triplets.
	For each triplet $ \bm{\tau}_i $ in $ \bm{T}_{out} $, we calculate the sum of the connection score regarding the samples in $ \mathcal{C} $ to greedily determine whether the samples in $ \bm{\tau}_i $ should be assigned into $ \mathcal{C} $ or not:
	\begin{equation}
	\begin{split}
	&\bm{\tau}^*=\arg\max_{\bm{\tau}} \ \sum_{j=1}^3\sum_\kappa^{|\bm{m}|} s_{\bm{n}_j\bm{m}_\kappa}, \\
	&s.t. \sum_{j=1}^3\sum_\kappa^{|\bm{m}|} s_{\bm{n}_j\bm{m}_\kappa}>1; \{\bm{x}_{\bm{n}_j}\}_{j=1}^3\in\bm{\tau};
	\{\bm{x}_{\bm{m}_\kappa}\}_{\kappa=1}^{|\bm{m}|}\in\mathcal{C},
	\end{split}
	\label{taustar}
	\end{equation}
	where $ \bm{n}, \bm{m} $ denote the set of indexes for the samples in $ \bm{\tau} $ and $\mathcal{C}$, respectively.
	We iteratively update the auxiliary sets $ \bm{T}_{out}^I $, $ \bm{T}_{in}^I $, $ \bm{X}_{out}^I $ and $ \bm{X}_{in}^I $ in the iterations.
	
	\vspace{1em}
	\noindent\textbf{Terminating:}
	We terminate the process of estimating the number of clusters and get $ \widetilde{K} $ clusters if and only if $ \bm{\tau}_{ini}^{\widetilde{K}+1}
	$ satisfies:
	\begin{equation}
	\rho(\bm{\tau}_{ini}^{\widetilde{K}+1},\bm{X}_{out}^{\widetilde{K}+1})\le\rho(\bm{\tau}_{ini}^{\widetilde{K}+1},\bm{X}_{in}^{\widetilde{K}+1}).
	\end{equation}
	Specifically, if the samples in $ \bm{\tau}_{ini}^{\widetilde{K}+1} $ are of high frequency in $ \bm{X}_{in}^{\widetilde{K}+1} $, \ie, the triplet with the highest local density in $ \bm{T}_{out}^{\widetilde{K}+1} $ is already contained in $ \bm{T}_{in}^{\widetilde{K}+1} $, we consider that the clusters are sufficient for modeling the intrinsic subspaces.
	
	\vspace{1em}
	\noindent\textbf{Avoiding Over-Segmentation:}
	We also introduce an alternative step to check the redundancy among initialized clusters $ \{\mathcal{C}_i\}_{i=1}^{\widetilde{K}} $ to avoid over-segmentation.
	We calculate the connection scores $ s $ for small-scale clusters against others,
	and merge the highly correlated clusters $ \mathcal{C}_i $ and $\mathcal{C}_j$ if we have
	\begin{equation}
	s_{ij}> \min (|\mathcal{C}_i|, |\mathcal{C}_j|),
	\label{sijcondition}
	\end{equation}
	where $|\mathcal{C}|$ denotes the number of samples in $\mathcal{C}$.
	We then get the initialized clusters $ \{\mathcal{C}_i\}_{i=1}^{\widehat{K}} $, where $ \widehat{K} $ is the estimated number of clusters and 
	~$\widehat{K}\le\widetilde{K}$.
	
	\begin{figure*}[t]
		\centering
		
		\includegraphics[width=0.9\textwidth]{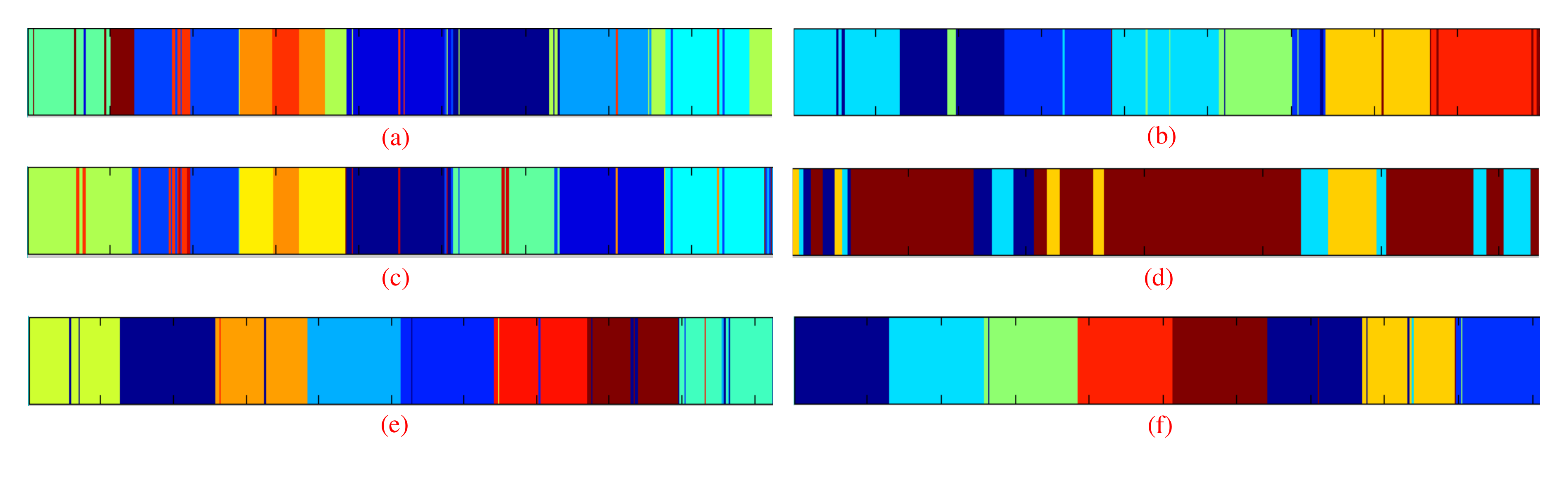}
		\vspace{-2em}
		\caption{Visualization of the clustering assignments for six methods, \ie, (a) SCAMS, (b) DP, (c) SVD, (d) DP-space, (e) autoSC-N and (f) autoSC.
			The experiments are conducted on the extended Yale B dataset with $8$ subjects.
			SCAMS over-segments the samples, while DP-space assigns the majority into one cluster.
			The proposed autoSC-N and auto-SC do not suffer from these problems.}
		\label{fig:clustering}
		
	\end{figure*}
	
	\vspace{1em}
	\noindent\textbf{Assigning Rest Samples:}
	%
	Given $ \{\mathcal{C}_i\}_{i=1}^{\widehat{K}} $, we assign each of the remaining samples into $\mathcal{C}$ which evokes an optimal fusion reward.
	For $ \bm{x}_j $, we find its optimal cluster $ \mathcal{C}^* $ by the following equation:
	\begin{equation}
	\mathcal{C}^*=\arg\max_{\mathcal{C}_i} \ R_f(\mathcal{C}_i|\bm{x}_j), \ \ i\in\{1,2,\cdots,\widehat{K}\},
	\label{rest}
	\end{equation}
	where $ R_f(\mathcal{C}|\bm{x}) $ is the fusion reward defined by~\eqref{fusion}.
	
	\subsection{An Extension: Neighboring based AutoSC Algorithm}
	
	\begin{algorithm}[t]
		\caption{\bf : Neighboring based autoSC (autoSC-N)}
		\label{algorithm_Neighbor}
		\hspace*{0.02in} {\bf Input:}
		$ \bm{X}=[\bm{x}_1,\cdots,\bm{x}_N]\in \mathbb{R}^{D\times N}$, $m$.
		\begin{algorithmic}[1]
			\State Normalize the magnitudes by $\bm{x}_i\leftarrow\bm{x}_i/\lVert \bm{x} \rVert_2$;
			\State Initialize the overall neighbor matrix $\mathcal{N}$ with $\mathcal{N} = \varnothing$;
			\For{$ j=1:N $}
			\State Initialize the spanned subspace $\mathcal{S}_j$ with $\mathcal{S}_j = \bm{x}_j$;
			\State Initialize the neighbor matrix $\mathcal{N}_j$ with $\mathcal{N}_j = \varnothing;$
			\For{$ I=1:m $}
			\State Calculate the similarity $s_{jk}$ between $\bm{x}_j$ and $ \bm{x}_k \in \bm{X}_{-\mathcal{N}}$ using~\eqref{s_jk};
			\State Update the spanned subspace $\mathcal{S}_j$ using~\eqref{S_update};
			\State $\mathcal{N}_j^{I+1} \leftarrow \mathcal{N}_j^I \ \cup\ \arg\max_{\bm{x}_k\in\bm{X}_{-\mathcal{N}}} s_{jk}$;
			\EndFor \State {\bf end for}
			\State $\mathcal{N} \leftarrow \mathcal{N} \cup \mathcal{N}_j$;
			\EndFor \State {\bf end for}
			\State Let $ \mathcal{N} $ replace the $ m $ Nearest Neighbors and conduct the steps from Step $ 6 $ to Step $ 21 $ in Algorithm~\ref{algorithm}.
		\end{algorithmic}
		\hspace*{0.02in} {\bf Output:}
		The cluster assignment $\{\mathcal{C}_i\}_{i=1}^{\widehat{K}}$.
	\end{algorithm}

	In Definition~\ref{mNN}, we collect $m$ nearest neighbors according to the magnitude of similarities between sample $ \bm{x}_j $ and all other samples in $\bm{X}_{-j}$.
	These similarities are depicted in $\bm{C}$ by optimizing~\eqref{intro} which is composed of a reconstruction loss term and a regularization term.
	In this subsection, we extend the autoSC with an alternative technique to find $ \bm{N}_m(\bm{x}_j)$ for each $\bm{x}_j$ based on greedy search.

	For each data sample $\bm{x}_j$, we let $ \mathcal{S}^I_j $ be the subspace spanned by $\bm{x}_j$ and its neighbors in the $I$-th iteration, where the neighbor set $ \mathcal{N}_j $ is initialized as $ \mathcal{N}^0_j = \varnothing $,
	and $ \mathcal{S}^I_j\in\mathbb{R}^{D\times\text{dim}(\mathcal{N}_j)} $.
	In each iteration, we measure the projected similarity between $ \bm{x}_j $ and other non-neighbor samples by calculating the orthonormal ordinates in the spanned subspace.
	For example, to calculate the similarity between $ \bm{x}_j $ and $ \bm{x}_k $ in the $I$-th iteration, we have
	\begin{equation}
	s_{jk} = \lVert (S^I_j)^\top\bm{x}_k \rVert_F^2,
	\label{s_jk}
	\end{equation}
	where $ \lVert \cdot \rVert_F^2 $ denotes the Frobenius norm and $ \bm{x}_k \in \bm{X}_{-\mathcal{N}}$.
	Consequently, for $ \bm{x}_j $ in the $I$-th iteration, we find the closest neighbor and update $ \mathcal{S}_j $ as follows:
	\begin{equation}
	\mathcal{S}_j^{I+1} \leftarrow \mathcal{S}_j^I \ \cup\ \arg\max_{\bm{x}_k\in\bm{X}_{-\mathcal{N}}} s_{jk}.
	\label{S_update}
	\end{equation}
	Here, we find one neighbor in each iteration and update the spanned subspace accordingly.
	The newly spanned subspace reflects more local structure of the ambient space which is assumed to cover the current sample.
	The neighbor set $\mathcal{N}_j$ is also updated by adding the new neighbor which is found in the $I$-th iteration.
	Finally, with $m$ iterations for each sample, we get an alternative $m$ nearest neighbor set $\mathcal{N}$.
	
	Given the neighbor matrix $\mathcal{N}$, we propose the neighboring based autoSC algorithm (autoSC-N) to directly discover the triplet relationship among data samples, followed by optimizing both model selection and fusion rewards for clustering.
	The main steps of autoSC-N are summarized in Algorithm~\ref{algorithm_Neighbor}.
	
	\subsection{Computational Complexity Analysis}
	
	In traditional subspace clustering system, the calculation of self-representation requires solving $N$ convex optimization problems over $D\times N$ constraints~\cite{Scalable_Sparse}.
	Spectral clustering is based on an eigen-decomposition operation on the Laplacian matrix followed by conducting K-means on the eigenvectors, both of which are time-consuming, involving a complex algebraic decomposition and iterative optimization, respectively~\cite{Belkin2001Laplacian, von2007tutorial}.
	The overall computational complexity can be more than $ O(N^3) $.
	For the proposed autoSC, it takes $O(Nm^2)$ to collect the triplet relationships for $N$ samples in the space spanned by the $m$ nearest neighbors.
	Here, since we have $ m\ll N $, the complexity of collecting the triplet relationships is $ O(N) $.
	The optimization of both model selection and fusion rewards takes $ O(Nn) $ where the number of triplets $ n $ has the same order of magnitude as $ N $.
	Specifically, we have $n\approx N\times\frac{m}{2}$, and thus the complexity of clustering is $O(N^2)$.
	
	For the extension, \ie, autoSC-N, collecting the neighbor matrix takes $ O(Nm) $ where the basic operation is a dot product of the $D$-dimensional tensors.
	This avoids the calculation of any convex optimization problem.

	\begin{table*}[t]
		\begin{center}{\scriptsize }
			\caption{Overall Comparison among comparative methods on subsets of the extended Yale B and COIL-20 datasets.
				The similarity matrix $ \bm{C} $ calculated by both SMR and LSR is utilized as the correlation matrix of SCAMS, DP, SVD and the proposed autoSC.
				The best results are in bold font while $^*$ indicates the second best performance.
				\label{table-overall}
			}
			\begin{tabular}{p{1cm}<{\centering}p{1.7cm}p{1cm}<{\centering}p{1cm}<{\centering}p{1cm}<{\centering}p{1cm}<{\centering}p{1cm}<{\centering}p{1cm}<{\centering}p{1cm}<{\centering}p{1cm}<{\centering}p{1cm}<{\centering}p{1cm}<{\centering}p{1cm}<{\centering}}
				\toprule
				\multirow{2}*{$\bm{C}$}&\multirow{2}*{Clustering}& \multirow{2}*{Metrics} & \multicolumn{5}{c}{extended Yale B}  & \multicolumn{4}{c}{COIL-20}\\
				\cmidrule(lr){4-8} \cmidrule(l){9-12}
				&&  & 8  & 15 &25 & 30&38 & 5  & 10 &15&20\\
				\midrule
				\multirow{2}*{LSR~\cite{Lu2012Robust}}&\multirow{2}*{SCAMS~\cite{Li2016Simultaneous,li2017simultaneous}}  
				& NC$ _e $                    & 5.21 & 14.00&17.12& 21.25&23.00&4.36 & 9.00& 18.32&21.00\\&
				& NMI                         & 0.1652 &   0.0643& 0.1544 &  0.3821&0.4236& 0.2435&0.1524 & 0.1124&0.1728\\
				\multirow{2}*{SMR~\cite{Hu2014Smooth}}&\multirow{2}*{SCAMS~\cite{Li2016Simultaneous,li2017simultaneous}}
				& NC$ _e $                    & 9.26 & 23.60&41.39 &76.22&81.00& 8.48& 19.72& 32.40&37.00\\
				&& NMI                         &  0.7183& 0.7272&0.6992& 0.7266&0.7425& 0.5885& 0.6527& 0.6668&0.6712 \\
				\midrule
				\multirow{2}*{LSR~\cite{Lu2012Robust}}&\multirow{2}*{DP~\cite{Rodriguez2014Clustering}}  
				& NC$ _e $                    & 7.90 & 98.38&127.92& 308.00&341.00& 10.90&   14.70& 301.05 &228.00\\&
				& NMI                         & 0.7060 & 0.6067&0.6245& 0.6516&0.6611&  0.7060  &  0.4984  &  0.6516 &0.5283\\
				\multirow{2}*{SMR~\cite{Hu2014Smooth}}&\multirow{2}*{DP~\cite{Rodriguez2014Clustering}}
				& NC$ _e $     &  3.06& 7.84&14.62& 24.76&29.00& 2.22& 5.30& 9.72 &11.00\\&
				& NMI & 0.6196& 0.5026&0.4391& 0.2166&0.2384& 0.6864& 0.4467& 0.3643&0.3547\\
				\midrule
				\multirow{2}*{LSR~\cite{Lu2012Robust}}&\multirow{2}*{SVD~\cite{Liu2013Robust}}  
				& NC$ _e $                    & 7.00 &9.42 &21.04&41.23 &44.00&2.76 &9.00 &12.05 &14.00\\&
				& NMI                         & 0.2412 &0.4304 &0.5567& 0.6523&0.6726& 0.6210& 0.1302&0.4092 &0.4125\\
				\multirow{2}*{SMR~\cite{Hu2014Smooth}}&\multirow{2}*{SVD~\cite{Liu2013Robust}}
				& NC$ _e $     & 2.40 & 9.06&11.65& 24.00&28.00& 0.48& 2.58& 8.36 &12.00\\&
				& NMI & 0.7078 & 0.4993&0.3739& 0.2808&0.2766& 0.7024& 0.7127& 0.7224 &0.7035\\
				\midrule
				\multirow{2}*{-}&\multirow{2}*{DP-space~\cite{Wang2015DP}}
				& NC$ _e $     & 2.08 & 8.96&15.75& 23.92&26.00& 0.78& 4.78& 9.38 &14.00\\&
				& NMI & 0.0343 & 0.0226&0.0432& 0.0406&0.0525& 0.0904& 0.0829& 0.0718 &0.0834\\
				\midrule
				\multirow{2}*{-}&\multirow{2}*{\textbf{autoSC-N}}
				& NC$ _e $ & 0.87$^*$& 3.16$^*$&4.32$^*$&7.68$^*$&9.00$^*$& 0.75$^*$& 2.21$^*$&2.42 &5.00\\&
				& NMI & 0.8306$^*$ & 0.7328&0.7165$^*$& 0.6566$^*$&0.6871$^*$& 0.7933$^*$&0.6216$^*$&0.7895$^*$ &0.7126\\
				\midrule
				\multirow{2}*{LSR~\cite{Lu2012Robust}}&\multirow{2}*{\textbf{autoSC}}  
				& NC$ _e $                    & {1.08} &  {3.32}&{5.79}&{10.50} &{12.00}&  {1.50}&     {3.40}&    {2.00}$^*$  & {4.00}$^*$\\&
				& NMI                         &  {0.8251} & {0.7375}$^*$ & {0.6871}& 0.5972&0.5833& {0.7786}& {0.5581} & \textbf{0.8670} & \textbf{0.8239}\\
				\multirow{2}*{SMR~\cite{Hu2014Smooth}}&\multirow{2}*{\textbf{autoSC}}
				& NC$ _e $     &\textbf{0.76} &\textbf{2.08}&\textbf{3.15}& \textbf{4.98}&\textbf{4.00}& \textbf{0.38}& \textbf{1.18}& \textbf{0.80}&\textbf{2.00}\\&
				& NMI & \textbf{0.9062} & \textbf{0.8589}&\textbf{0.8432}& \textbf{0.8287}&\textbf{0.7943}& \textbf{0.8315}& \textbf{0.7701}&  {0.7266} & {0.7568}$^*$\\
				
				\bottomrule
			\end{tabular}
			
		\end{center}
		
	\end{table*}

	\section{Experiments}
	
	\subsection{Experimental Setup}
	
	In the experiments, we compare the 
	automatic methods on the benchmark datasets, \ie, the extended Yale B~\cite{Georghiades2001From} and the COIL-20~\cite{Rate2011Columbia} dataset, followed by verifying the robustness of the proposed method
	to different $ \bm{C} $ derived from various self-representation schemes
	along with combinations of different methods for estimating the number of clusters and segmenting the samples.
	We design comprehensive evaluation metrics to validate the clustering performance, \ie, the error rate of the number of clusters and the triplets.
	For all experiments on subsets, the reported results are the average of $50$ trials.
	We also conduct experiments on a motion segmentation task using the Hopkins 155 dataset.
	
	\subsubsection{Datasets}
	
	The extended Yale B~\cite{Georghiades2001From} dataset is a widely used face clustering dataset which contains face images with different illumination of $ 38 $ subjects, each subject has $ 64 $ images.
	
	The COIL-20~\cite{Rate2011Columbia} dataset consists of $ 20 $ different real subjects, including cups, bottles and so on.
	For each subject, there are $ 72 $ images with different camera viewpoints.

	The Hopkins 155 dataset~\cite{tron2007benchmark} consists of $ 155 $ video sequences. 
	For each video sequence, there are $ 2 $ or $ 3 $ motions.
	
	\subsubsection{Comparative Methods}
	We make comparisons with the following methods: SCAMS~\cite{Li2016Simultaneous, li2017simultaneous}, a density peak based method (DP)~\cite{Rodriguez2014Clustering}, a singular value decomposition based method (SVD)~\cite{Liu2013Robust} and DP-space~\cite{Wang2015DP}.
	Besides, we utilize the following subspace representation methods to generate different coefficient matrices $ \bm{C} $: LRR~\cite{Liu2010Robust}, CASS~\cite{Lu2015Correlation}, LSR~\cite{Lu2012Robust}, SMR~\cite{Hu2014Smooth} and ORGEN~\cite{You2016Oracle}.
	The similarity matrix $ \bm{C} $ is then used to calculate the triplet relationships for autoSC.
	
	\subsubsection{Evaluation Metrics} To evaluate the performance of the proposed triplets, we define the error rate $ \mathcal{A} $ as follows:
	\begin{equation}\label{accoftri}
	\mathcal{A}=\frac{1}{n}\sum_{i=1}^{n}\frac{3-\sigma(\bm{\tau}_i|\bm{g}_i^*)}{2},
	\end{equation}
	where $ n $ denotes the number of the triplets and $ \sigma(\bm{\tau}|\bm{g}^*) $ is the counting function on the frequency that $ \bm{x}\in\bm{g}^*$ for all $ \bm{x}\in\bm{\tau} $.
	Here the output of $ f $ ranges from $ 0 $ to $ 2 $.
	The dynamic set $ \bm{g}_i^* $ consists of samples in one subspace $ S $ according to the ground truth, where $ S $ contains as many samples in $ \bm{\tau}_i $ as possible.
	
	We introduce the error rate of the number of clusters (NC$_e$) as the primary evaluation metric for the clustering methods which estimate the number of clusters $ \widehat{K} $ automatically:
	\begin{equation}
	\text{NC}_e = \frac{1}{M}\sum_{i=1}^M|\widehat{K}_i-K|,
	\end{equation}
	where $K$ is the real number of clusters, $ M $ is the number of trials and $ \widehat{K}_i $ is the estimated number of clusters in the $ i $-th trial.
	We also use the standard normalized mutual information (NMI)~\cite{Li2003Clustering} to measure the similarity between two clustering distributions, \ie, the prediction and the ground truth.
	With respect to NMI, the entropy illustrates the nondeterminacy of one clustering to the other, and the mutual information quantifies the amount of information that one variable obtains from the other.

	\subsubsection{Parameter $m$}
	The parameter $m$ in Definition~\ref{mNN}, \ie, the number of preserved neighbors for each sample, is related to the intrinsic dimension of the subspaces.
	We empirically evaluate the influence of $m$ on both extended Yale B and COIL-20 datasets with 15 subjects.
	Besides, we use subspace representations derived from SMR.
	The results are shown in Table~\ref{evaluate_m}.
	As shown in the table, the proposed method achieves best performance when we have $m=8$ for most cases.
	Actually, the parameter $m$ is robust since the performance is stable when $m>8$.

	\begin{table}[t]
		\begin{center}
			{\scriptsize }
			\caption{Clustering performance of the proposed autoSC with different $m$ on both extended Yale B (EYaleB) and COIL-20 (Coil-20) datasets with 
			15 subjects.
			The subspace representation is derived from SMR.
			Based on the results, we set $m=8$ in the rest of the paper.
			}
			\label{evaluate_m}
			\begin{tabular}{m{0.6cm}<{\centering}m{0.6cm}<{\centering}m{0.6cm}<{\centering}m{0.5cm}<{\centering}m{0.5cm}<{\centering}m{0.5cm}<{\centering}m{0.5cm}<{\centering}m{0.5cm}<{\centering}m{0.5cm}<{\centering}}
				\toprule
				$m$& Metrics &5 & 6 & 7 & 8 & 9& 10 & 11\\
				\midrule
				\multirow{2}*{EYaleB}&NC$ _e $ &4.52&3.68&3.13&\textbf{2.08}&2.12&2.29&2.18\\
				&NMI &0.7125&0.7736&0.8047&\textbf{0.858}&0.8551&0.8423&0.8536\\
				\multirow{2}*{Coil-20}&NC$ _e $ &1.68&1.29&0.92&0.80&0.88&\textbf{0.76}&0.98\\
				&NMI&0.5647&0.6157&0.6774&\textbf{0.7266}&0.7107&0.7211&0.7120\\
				\bottomrule
			\end{tabular}
		\end{center}
	\end{table}

	\begin{figure}[t]
		\centering
		\includegraphics[width=0.48\textwidth]{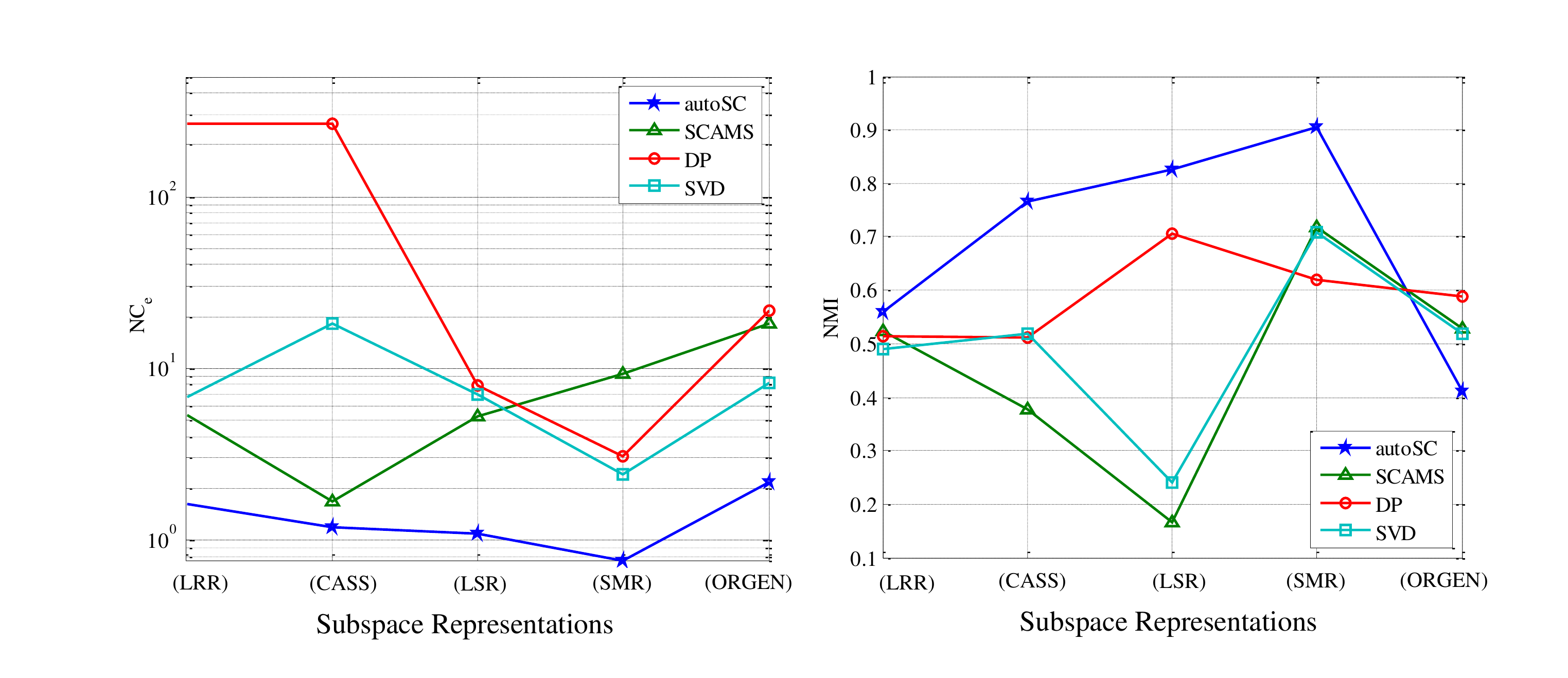}
		\caption{Clustering results using different self-representation schemes on the extended Yale B dataset with 8 subjects.
			The left figure denotes the comparison of four methods using the NC$_e$ metric while the right one uses NMI.
			Each point in the curves is derived by the combination of different clustering methods and self-representation schemes.
			The proposed autoSC achieves consistent performance on the evaluation of NC$_e$.
			\label{extension}}
	\end{figure}
	
	\begin{table*}[t]
		\begin{center}{\scriptsize }
			\caption{Error rates of triplets ($\mathcal{A}$) for the proposed autoSC on 
			subsets of the extended Yale B (eYaleB) and COIL-20 datasets.
				For each row, we utilize the similarity matrix $ \bm{C} $ derived from one self-representation method in the first column.
				The autoSC achieves consistency on the calculation of the triplets.
			}
			\label{table-acctri}
			\begin{tabular}{p{1.6cm}<{\centering}p{1.2cm}<{\centering}p{1.2cm}<{\centering}p{1.2cm}<{\centering}p{1.2cm}<{\centering}p{1.2cm}<{\centering}p{1.2cm}<{\centering}p{1.2cm}<{\centering}p{1.2cm}<{\centering}p{1.2cm}<{\centering}p{1.2cm}<{\centering}}
				\toprule
				\multirow{2}*{$\bm{C}$} &\multicolumn{5}{c}{extended Yale B}  & \multicolumn{4}{c}{COIL-20}\\
				\cmidrule(lr){2-6} \cmidrule(l){7-10}
				& 8  & 15 &25 & 30&38 & 5  & 10 &15&20\\
				\midrule
				LRR~\cite{Liu2010Robust}
				& 0.0155&0.0147 &0.0158 &0.0176&0.0169&0.0185&0.0252 &0.0224 &0.0231\\
				CASS~\cite{Lu2015Correlation}
				& 0.0158&0.0148 & 0.0140&0.0157&0.0162&0.0195 &0.0198 &0.0203 &0.0193\\
				LSR~\cite{Lu2012Robust}
				& 0.0144& 0.0148& 0.0162&0.0181&0.0172&0.0188& 0.0188& 0.0212&0.0199\\
				SMR~\cite{Hu2014Smooth}
				& 0.0135&0.0149 & 0.0154&0.0181&0.0161&0.0175 & 0.0182&0.0196 &0.0202\\
				ORGEN~\cite{You2016Oracle}
				& 0.0166&0.0145 &0.0151 &0.0177&0.0169& 0.0196& 0.0210&0.0215 &0.0220\\
				\bottomrule
			\end{tabular}
			
		\end{center}
		
	\end{table*}
	
	\begin{figure*}[t]
		\centering
		
		\includegraphics[width=0.8\textwidth]{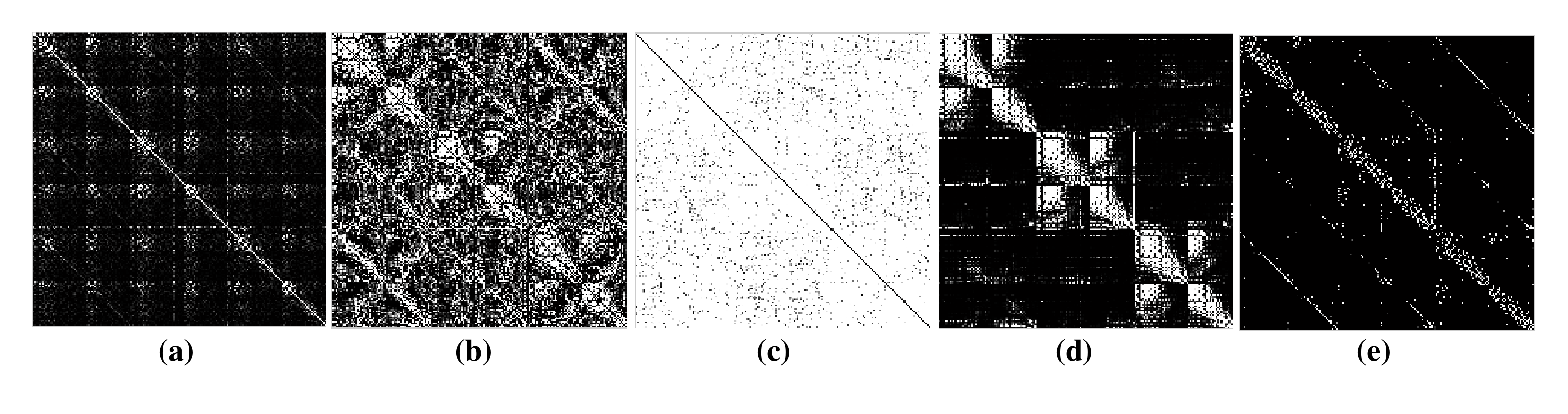}
		\vspace{-2em}
		\caption{Visualization of the similarity matrix $ \bm{C} $ derived from different self-representation schemes on the extended Yale B dataset with $3$ subjects.
			The white regions denote the locations with non-zero coefficients.
			Different methods, \ie, (a) LRR, (b) CASS, (c) LSR, (d) SMR and (e) ORGEN, produce $ \bm{C} $ with different characteristics, \eg, (d) derived from SMR is block-diagonal, while (e) derived from ORGEN is sparse.
			\label{fig:Ws}}
	\end{figure*}
	
	\subsection{Comparisons among Automatic Clustering}
	
	We conduct experiments on the extended Yale B and COIL-20 datasets with different numbers of subjects, and compare four methods with the proposed autoSC and autoSC-N on the metrics of NC$ _e $ and NMI.
	For SCAMS~\cite{Li2016Simultaneous, li2017simultaneous}, DP~\cite{Rodriguez2014Clustering}, SVD~\cite{Liu2013Robust} and our autoSC, the optimization module in SMR~\cite{Hu2014Smooth} is employed to generate the similarity matrix $ \bm{C} $.
	The DP-space method simultaneously estimates $ \widehat{K} $ and finds the subspaces without the requirement of a similarity matrix.
	All parameters of the contrasted methods are tuned to provide the best performance.
	
	Fig.~\ref{fig:clustering} and Table~\ref{table-overall} report the performance.
	As shown in Table~\ref{table-overall}, when combining SMR, the averaged NC$ _e $ of autoSC is smaller than other comparative methods on all experimental configurations, indicating that it gives a close estimation on the number of clusters.
	For example, the estimated $ \widehat{K} $ on the extended Yale B with $ 8 $ subjects has a deviation of less than $ 1 $, and produces a NMI higher than $ 0.9 $.
	The autoSC-N gets the second best performance on most configurations, which demonstrates the effectiveness of both triplet relationship and reward optimization.
	In contrast, SVD achieves comparable results on the small-scale configuration of each dataset, but the performance becomes poor when the number of samples increases.
	It is mainly because the largest gap between the pair of 
	singular values decreases when the number of clusters becomes larger.
	When combining SMR, SCAMS performs comparably according to NMI on both datasets, however, as is illustrated in Fig.~\ref{fig:clustering} (a) and Table~\ref{table-overall}, it provides a much larger $ \widehat{K} $ than the ground truth, \eg, $ \widehat{K}>100 $ when $ K=30 $ on the extended Yale B dataset.
	NMI does not
	strongly
	penalize over-segmentation, making the metric NC$ _e $ be the primary evaluation of the SCAMS method.
	The DP-space performs well on NC$ _e $, but has poor performance on the NMI.
	This is because most samples are assigned into one cluster, and the other clusters are small.
	In addition, when combining LSR, as shown in Table~\ref{table-overall}, the performance of all methods decrease, while the proposed autoSC still achieves best performance on most configurations.
	It demonstrates the generalization ability of our autoSC.

	\subsection{Robustness to Self-Representations Schemes}
	
	The methods including SCAMS~\cite{Li2016Simultaneous, li2017simultaneous}, DP~\cite{Rodriguez2014Clustering}, SVD~\cite{Liu2013Robust} and the proposed autoSC require the similarity matrix $ \bm{C}  $ as input.
	Also, for DP~\cite{Rodriguez2014Clustering}, the distance among samples needs to be calculated.
	We calculate the distance $ d_{ij} $ between samples $ \bm{x}_i $ and $ \bm{x}_j $ by $ d_{ij}=\frac{1}{c_{ij}} $ rather than the simple Euclidean distance.
	To verify the robustness of the proposed autoSC regarding various subspace representations, we calculate the similarity matrix $ \bm{C}  $ using $ 5 $ subspace representation modules, followed by the combinations with the $ 4 $ methods which automatically estimate the number of clusters and segment the samples.
	
	Table~\ref{table-acctri} shows the evaluation results of $\mathcal{A}$ on both datasets with the combinations of $5$ subspace representations, while the NC$_e$ and NMI on the extended Yale B dataset with $ 8 $ subjects are reported in Fig.~\ref{extension}.
	Moreover, we visualize the similarity matrix $ \bm{C} $ derived from $5 $ subspace representation modules in Fig.~\ref{fig:Ws}.
	We can see from Fig.~\ref{extension} that the SCAMS, DP and SVD methods are sensitive to the choice of the subspace representation module.
	For example, DP estimates $ \widehat{K} $ as a relatively close value to the ground truth when combined with SMR (NC$ _e =3.06$), but generates a totally wrong estimation when combined with LRR (NC$ _e =265.60$).
	Different subspace representation modules generate coefficient matrices with various intrinsic properties~\cite{Vidal2010A}, thus the parameter for truncation error $ \epsilon $ needs to be tuned carefully.
	
	For the proposed autoSC, it is stable on different combinations considering the metric of NC$ _e $ and $ \mathcal{A} $, which demonstrates the complementary ability of the proposed method.
	For all combinations, the error rate of the triplets obtained from~\eqref{tau_calculation} is less than $ 2\% $, which guarantees the consistency of the proposed autoSC with different kinds of $ \bm{C} $.
	Furthermore, it shows better performance when combined with CASS, LSR and SMR than other combinations on both metrics in Fig.~\ref{extension}.
	The reason lies on the guarantee of the mapping invariance which is termed as the grouping effect~\cite{Lu2015Correlation,Lu2012Robust,Hu2014Smooth}, together with the filtering
	of weak connections and the self-constraint among samples within
	triplets.
	As shown in Fig.~\ref{fig:Ws} (b), (c), (d), the coefficient matrices are dense while it shows block-diagonal structure in Fig.~\ref{fig:Ws} (d) and each block corresponds to one cluster.
	Therefore, the nearest neighbors which are used to generate the triplets can be chosen precisely.
	The performance decreases when combined with ORGEN since the similarity matrix $ \bm{C} $ derived from ORGEN is sparse with less 
	locations for constructing effective triplets.
	
	\begin{table}[t]
		\begin{center}
			{\scriptsize}
			\caption{Comparison of time-consumption (in seconds) on the extended Yale B dataset.
				The best results are in bold font while $^*$ indicates the second best performance.\label{time}
			}
			\begin{tabular}{m{0.8cm}<{\centering}m{0.9cm}<{\centering}m{0.7cm}<{\centering}m{0.7cm}<{\centering}m{0.9cm}<{\centering}m{0.9cm}<{\centering}m{0.9cm}<{\centering}}
				\toprule
				Subjects& SCAMS   & DP  & SVD   & DP-space&autoSC-N & autoSC\\
				\midrule
				8&12.45&6.92&17.32&9.52&\textbf{1.69}&2.02$^*$\\
				15&30.12&13.04&44.13&18.05&\textbf{4.72}&5.79$^*$\\
				25&125.66&33.76&146.78&36.94&\textbf{10.28}&19.81$^*$\\
				30&175.80&59.02&225.93&67.89&\textbf{16.33}&36.06$^*$\\
				38&267.07&97.69&314.28&104.50&\textbf{29.45}&62.35$^*$\\
				\bottomrule
			\end{tabular}
		\end{center}
	\end{table}
	\subsection{Time Efficiency}
	
	Table~\ref{time} shows the run-time of comparative methods using subsets from the extended Yale B dataset.
	The experiments are conducted on a machine with a $ 2.93 $GHz CPU and $ 32 $GB RAM.
	AutoSC-N requires the least run-time compared to all comparative methods due to the following two reasons.
	First, autoSC-N explores the neighborhood relationship in the raw data space rather than solving a convex optimization problem.
	Second, it employs a greedy optimization scheme to estimate the number of clusters and calculate the clustering assignment rather than a complex optimizing method such as computing the singular value decomposition.
	Note the proposed autoSC method achieves the second best result among comparative methods.
	
	\begin{table}[t]
		\begin{center}
			{\scriptsize }
			\caption{Motion segmentation performance on Hopkins 155 dataset.
				The row of `Time' reports the average time-consumption on handling each video sequence.
				The best results are in bold font while $^*$ indicates the second best performance.
				The proposed autoSC outperforms other comparative methods regarding both the estimation of $K$ and clustering.
			}
			\label{motionSeg}
			\begin{tabular}{m{0.8cm}<{\centering}m{0.9cm}<{\centering}m{0.7cm}<{\centering}m{0.7cm}<{\centering}m{0.9cm}<{\centering}m{0.9cm}<{\centering}m{0.9cm}<{\centering}}
				\toprule
				Metrics& SCAMS   & DP  & SVD   & DP-space&autoSC-N & autoSC\\
				\midrule
				$\text{NC}_e$&3.67&2.12&1.29&2.97&0.52$^*$&\textbf{0.18}\\
				NMI &0.7892&0.8233&0.8670&0.7921&0.9155$^*$&\textbf{0.9871}\\
				Time$ /s $&2.68&1.22&2.45&1.56&\textbf{0.26}&0.32$^*$\\
				\bottomrule
			\end{tabular}
		\end{center}
	\end{table}
	
	\subsection{Real Application: Motion segmentation}

	Motion segmentation refers to the task of segmenting multiple video sequence.
	The candidate video is composed of multiple foreground objects, which are rapidly moving and required to be clustered into spatiotemporal regions corresponding to specific motions.
	Following the traditional scheme~\cite{elhamifar2013sparse}, we consider the Hopkins 155 dataset~\cite{tron2007benchmark} and solve the motion segmentation problem by first extracting a set of feature points for each frame followed by clustering them based on the motions.
	Table~\ref{motionSeg} reports the comparison against four automatic clustering methods.
	For SCAMS~\cite{Li2016Simultaneous, li2017simultaneous}, DP~\cite{Rodriguez2014Clustering} and SVD~\cite{Liu2013Robust}, the SMR~\cite{Hu2014Smooth} is firstly conducted to calculate the similarity matrix.
	As shown in the table, the proposed autoSC achieves best performance on both metrics, indicating that the autoSC is effective at both estimating the number of motions (about $0.18$ error rate) and segmenting the feature points (obtains NMI of more than $ 0.98 $).
	In addition, it shows favorable efficiency on the motion segmentation task.
	The autoSC-N is the most efficient method ($ 0.26s $ per sequence) with second best performance on NC$_e$ and NMI.
	The SVD method obtains the best result among other comparative methods, but it consumes much more time (about more than $ 2.5s $ per sequence) due to the singular value decomposition process.

	\section{Conclusion}

	In this paper, we propose a joint model to estimate the number of clusters and segment the samples in a data set.
	Based on the self-representation of dataset, we first design a hyper-correlation oriented meta-element termed as the triplet relationship, which indicates a compact local structure among three samples.
	The triplet is more robust than pairwise relationships when partitioning samples near the intersection of two subspaces due to the complementarity of mutual restrictions.
	Accordingly, we propose the autoSC method to optimize two reward functions simultaneously, of which the model selection reward constrains the number of clusters and the fusion reward facilitates the clustering assignment of the samples.
	Both functions are greedily maximized during the clustering process.
	In addition, we provide an extension of autoSC which automatically calculates the neighboring relationship in the raw data space rather than a similarity space spanned by self-representation.
	Experimental results on face clustering, synthetic dataset clustering and motion segmentation tasks demonstrate the effectiveness and efficiency of our approaches.

	\bibliographystyle{IEEEtran}
	\bibliography{TIP_autoSC}

\begin{thebibliography}{10}
\providecommand{\url}[1]{#1}
\csname url@samestyle\endcsname
\providecommand{\newblock}{\relax}
\providecommand{\bibinfo}[2]{#2}
\providecommand{\BIBentrySTDinterwordspacing}{\spaceskip=0pt\relax}
\providecommand{\BIBentryALTinterwordstretchfactor}{4}
\providecommand{\BIBentryALTinterwordspacing}{\spaceskip=\fontdimen2\font plus
\BIBentryALTinterwordstretchfactor\fontdimen3\font minus
  \fontdimen4\font\relax}
\providecommand{\BIBforeignlanguage}[2]{{%
\expandafter\ifx\csname l@#1\endcsname\relax
\typeout{** WARNING: IEEEtran.bst: No hyphenation pattern has been}%
\typeout{** loaded for the language `#1'. Using the pattern for}%
\typeout{** the default language instead.}%
\else
\language=\csname l@#1\endcsname
\fi
#2}}
\providecommand{\BIBdecl}{\relax}
\BIBdecl

\bibitem{yang2018AAAI}
J.~Yang, J.~Liang, K.~Wang, Y.-L. Yang, and M.-M. Cheng, ``Automatic model
  selection in subspace clustering via triplet relationships,'' in \emph{AAAI
  Conference on Artificial Intelligence}, 2018.

\bibitem{wang2017exclusivity}
X.~Wang, X.~Guo, Z.~Lei, C.~Zhang, and S.~Z. Li, ``Exclusivity-consistency
  regularized multi-view subspace clustering,'' in \emph{IEEE Conference on
  Computer Vision and Pattern Recognition}, 2017, pp. 923--931.

\bibitem{peng2017constructing}
X.~Peng, Z.~Yu, Z.~Yi, and H.~Tang, ``Constructing the {L2-graph} for robust
  subspace learning and subspace clustering,'' \emph{IEEE Transactions on
  Cybernetics}, vol.~47, no.~4, pp. 1053--1066, 2017.

\bibitem{Vidal2010A}
R.~Vidal, ``Subspace clustering,'' \emph{IEEE Signal Processing Magazine},
  vol.~28, no.~2, pp. 52--68, 2011.

\bibitem{jia2017subspace}
H.~Jia and Y.-M. Cheung, ``Subspace clustering of categorical and numerical
  data with an unknown number of clusters,'' \emph{IEEE Transactions on Neural
  Networks and Learning Systems}, vol.~29, no.~8, pp. 3308--3325, 2018.

\bibitem{xu2014clustering}
X.~Xu, Z.~Huang, D.~Graves, and W.~Pedrycz, ``A clustering-based graph
  laplacian framework for value function approximation in reinforcement
  learning,'' \emph{IEEE Transactions on Cybernetics}, vol.~44, no.~12, pp.
  2613--2625, 2014.

\bibitem{zhu2017subspace}
P.~Zhu, W.~Zhu, Q.~Hu, C.~Zhang, and W.~Zuo, ``Subspace clustering guided
  unsupervised feature selection,'' \emph{Pattern Recognition}, vol.~66, pp.
  364--374, 2017.

\bibitem{cao2015constrained}
X.~Cao, C.~Zhang, C.~Zhou, H.~Fu, and H.~Foroosh, ``Constrained multi-view
  video face clustering,'' \emph{IEEE Transactions on Image Processing},
  vol.~24, no.~11, pp. 4381--4393, 2015.

\bibitem{elhamifar2009sparse}
E.~Elhamifar and R.~Vidal, ``Sparse subspace clustering,'' in \emph{IEEE
  Conference on Computer Vision and Pattern Recognition}, 2009, pp. 2790--2797.

\bibitem{yangl0}
Y.~Yang, J.~Feng, N.~Jojic, J.~Yang, and T.~S. Huang, ``$ l_0 $-sparse subspace
  clustering,'' in \emph{European Conference on Computer Vision}, 2016, pp.
  731--747.

\bibitem{shi2000normalized}
J.~Shi and J.~Malik, ``Normalized cuts and image segmentation,'' \emph{IEEE
  Transactions on Pattern Analysis and Machine Intelligence}, vol.~22, no.~8,
  pp. 888--905, 2000.

\bibitem{Wang2015DP}
Y.~Wang and J.~Zhu, ``{DP-space}: Bayesian nonparametric subspace clustering
  with small-variance asymptotics,'' in \emph{International Conference on
  Machine Learning}, 2015, pp. 862--870.

\bibitem{Li2016Simultaneous}
Z.~Li, S.~Yang, L.~F. Cheong, and K.~C. Toh, ``Simultaneous clustering and
  model selection for tensor affinities,'' in \emph{IEEE Conference on Computer
  Vision and Pattern Recognition}, 2016, pp. 5347--5355.

\bibitem{javed2017background}
S.~Javed, A.~Mahmood, T.~Bouwmans, and S.~K. Jung, ``Background-foreground
  modeling based on spatiotemporal sparse subspace clustering,'' \emph{IEEE
  Transactions on Image Processing}, vol.~26, no.~12, pp. 5840--5854, 2017.

\bibitem{kumar2016hybrid}
D.~Kumar, J.~C. Bezdek, M.~Palaniswami, S.~Rajasegarar, C.~Leckie, and T.~C.
  Havens, ``A hybrid approach to clustering in big data,'' \emph{IEEE
  Transactions on Cybernetics}, vol.~46, no.~10, pp. 2372--2385, 2016.

\bibitem{elhamifar2013sparse}
E.~Elhamifar and R.~Vidal, ``Sparse subspace clustering: Algorithm, theory, and
  applications,'' \emph{IEEE Transactions on Pattern Analysis and Machine
  Intelligence}, vol.~35, no.~11, pp. 2765--2781, 2013.

\bibitem{Nasihatkon2011Graph}
B.~Nasihatkon and R.~Hartley, ``Graph connectivity in sparse subspace
  clustering,'' in \emph{IEEE Conference on Computer Vision and Pattern
  Recognition}, 2011, pp. 2137--2144.

\bibitem{zhan2018graph}
K.~Zhan, C.~Zhang, J.~Guan, and J.~Wang, ``Graph learning for multiview
  clustering,'' \emph{IEEE transactions on cybernetics}, no.~99, pp. 1--9,
  2017.

\bibitem{Rodriguez2014Clustering}
A.~Rodriguez and A.~Laio, ``Clustering by fast search and find of density
  peaks,'' \emph{Science}, vol. 344, no. 6191, pp. 1492--1496, 2014.

\bibitem{Scalable_Sparse}
X.~Peng, L.~Zhang, and Z.~Yi, ``Scalable sparse subspace clustering,'' in
  \emph{IEEE Conference on Computer Vision and Pattern Recognition}, 2013, pp.
  430--437.

\bibitem{Gao2013Laplacian}
S.~Gao, I.~W. Tsang, and L.~T. Chia, ``Laplacian sparse coding, hypergraph
  laplacian sparse coding, and applications,'' \emph{IEEE Transactions on
  Pattern Analysis and Machine Intelligence}, vol.~35, no.~1, pp. 92--104,
  2013.

\bibitem{Kim2014Image}
S.~Kim, D.~Y. Chang, S.~Nowozin, and P.~Kohli, ``Image segmentation
  usinghigher-order correlation clustering,'' \emph{IEEE Transactions on
  Pattern Analysis and Machine Intelligence}, vol.~36, no.~9, pp. 1761--1774,
  2014.

\bibitem{schroff2015facenet}
F.~Schroff, D.~Kalenichenko, and J.~Philbin, ``Facenet: A unified embedding for
  face recognition and clustering,'' in \emph{International Conference on
  Computer Vision}, 2015, pp. 815--823.

\bibitem{wang2014constraint}
H.~Wang, T.~Li, T.~Li, and Y.~Yang, ``Constraint neighborhood projections for
  semi-supervised clustering,'' \emph{IEEE Transactions on Cybernetics},
  vol.~44, no.~5, pp. 636--643, 2014.

\bibitem{li2017structured}
C.-G. Li, C.~You, and R.~Vidal, ``Structured sparse subspace clustering: A
  joint affinity learning and subspace clustering framework,'' \emph{IEEE
  Transactions on Image Processing}, vol.~26, no.~6, pp. 2988--3001, 2017.

\bibitem{zhang2017flexible}
C.~Zhang, H.~Fu, Q.~Hu, P.~Zhu, and X.~Cao, ``Flexible multi-view
  dimensionality co-reduction,'' \emph{IEEE Transactions on Image Processing},
  vol.~26, no.~2, pp. 648--659, 2017.

\bibitem{you2016scalable}
C.~You, D.~Robinson, and R.~Vidal, ``Scalable sparse subspace clustering by
  orthogonal matching pursuit,'' in \emph{IEEE Conference on Computer Vision
  and Pattern Recognition}, 2016, pp. 3918--3927.

\bibitem{Cheng_2016_CVPR}
Y.~Cheng, Y.~Wang, M.~Sznaier, and O.~Camps, ``Subspace clustering with priors
  via sparse quadratically constrained quadratic programming,'' in \emph{IEEE
  Conference on Computer Vision and Pattern Recognition}, 2016, pp. 5204--5212.

\bibitem{li2017simultaneous}
Z.~Li, L.-F. Cheong, S.~Yang, and K.-C. Toh, ``Simultaneous clustering and
  model selection: Algorithm, theory and applications,'' \emph{IEEE
  Transactions on Pattern Analysis and Machine Intelligence}, vol.~40, no.~8,
  pp. 1964--1978, 2018.

\bibitem{wu2016ordered}
F.~Wu, Y.~Hu, J.~Gao, Y.~Sun, and B.~Yin, ``Ordered subspace clustering with
  block-diagonal priors,'' \emph{IEEE Transactions on Cybernetics}, vol.~46,
  no.~12, pp. 3209--3219, 2016.

\bibitem{Li2015Structured}
C.~G. Li and R.~Vidal, ``Structured sparse subspace clustering: A unified
  optimization framework,'' in \emph{IEEE Conference on Computer Vision and
  Pattern Recognition}, 2015, pp. 277--286.

\bibitem{Hu2014Smooth}
H.~Hu, Z.~Lin, J.~Feng, and J.~Zhou, ``Smooth representation clustering,'' in
  \emph{IEEE Conference on Computer Vision and Pattern Recognition}, 2014, pp.
  3834--3841.

\bibitem{You2016Oracle}
C.~You, C.~G. Li, D.~P. Robinson, and R.~Vidal, ``Oracle based active set
  algorithm for scalable elastic net subspace clustering,'' in \emph{IEEE
  Conference on Computer Vision and Pattern Recognition}, 2016, pp. 3928--3937.

\bibitem{fang2016robust}
X.~Fang, Y.~Xu, X.~Li, Z.~Lai, and W.~K. Wong, ``Robust semi-supervised
  subspace clustering via non-negative low-rank representation,'' \emph{IEEE
  Transactions on Cybernetics}, vol.~46, no.~8, pp. 1828--1838, 2016.

\bibitem{rahmani2017innovation}
M.~Rahmani and G.~Atia, ``Innovation pursuit: A new approach to the subspace
  clustering problem,'' in \emph{International Conference on Machine Learning},
  2017, pp. 2874--2882.

\bibitem{Dyer2013Greedy}
E.-L. Dyer, A.-C. Sankaranarayanan, and R.-G. Baraniuk, ``Greedy feature
  selection for subspace clustering,'' \emph{Journal of Machine Learning
  Research}, vol.~14, no.~1, pp. 2487--2517, 2013.

\bibitem{park2014greedy}
D.~Park, C.~Caramanis, and S.~Sanghavi, ``Greedy subspace clustering,'' in
  \emph{Advances in Neural Information Processing Systems}, 2014, pp.
  2753--2761.

\bibitem{purkait2017clustering}
P.~Purkait, T.-J. Chin, A.~Sadri, and D.~Suter, ``Clustering with hypergraphs:
  the case for large hyperedges,'' \emph{IEEE Transactions on Pattern Analysis
  and Machine Intelligence}, vol.~39, no.~9, pp. 1697--1711, 2017.

\bibitem{Liu2010Robust}
H.~Liu, L.~Latecki, and S.~Yan, ``Robust clustering as ensembles of affinity
  relations,'' in \emph{Advances in Neural Information Processing Systems},
  2010, pp. 1414--1422.

\bibitem{li2016structured}
C.-G. Li and R.~Vidal, ``A structured sparse plus structured low-rank framework
  for subspace clustering and completion,'' \emph{IEEE Transactions on Signal
  Processing}, vol.~64, no.~24, pp. 6557--6570, 2016.

\bibitem{guo2014spatial}
Y.~Guo, J.~Gao, and F.~Li, ``Spatial subspace clustering for drill hole
  spectral data,'' \emph{Journal of Applied Remote Sensing}, vol.~8, no.~1, p.
  083644, 2014.

\bibitem{Feng2014Robust}
J.~Feng, Z.~Lin, H.~Xu, and S.~Yan, ``Robust subspace segmentation with
  block-diagonal prior,'' in \emph{IEEE Conference on Computer Vision and
  Pattern Recognition}, 2014, pp. 3818--3825.

\bibitem{wang2016product}
B.~Wang, Y.~Hu, J.~Gao, Y.~Sun, and B.~Yin, ``Product {Gr}assmann manifold
  representation and its \protect{LRR} models,'' in \emph{AAAI Conference on
  Artificial Intelligence}, 2016, pp. 2122--2129.

\bibitem{liu2016decentralized}
B.~Liu, X.-T. Yuan, Y.~Yu, Q.~Liu, and D.-N. Metaxas, ``Decentralized robust
  subspace clustering,'' in \emph{AAAI Conference on Artificial Intelligence},
  2016, pp. 3539--3545.

\bibitem{Xiao2015FaLRR}
S.~Xiao, W.~Li, D.~Xu, and D.~Tao, ``\protect{FaLRR}: A fast low rank
  representation solver,'' in \emph{IEEE Conference on Computer Vision and
  Pattern Recognition}, 2015, pp. 4612--4620.

\bibitem{Lu2012Robust}
C.~Y. Lu, H.~Min, Z.~Q. Zhao, L.~Zhu, D.~S. Huang, and S.~Yan, ``Robust and
  efficient subspace segmentation via least squares regression,'' in
  \emph{European Conference on Computer Vision}, 2012, pp. 347--360.

\bibitem{liu2016deterministic}
G.~Liu, H.~Xu, J.~Tang, Q.~Liu, and S.~Yan, ``A deterministic analysis for
  {LRR},'' \emph{IEEE Transactions on Pattern Analysis and Machine
  Intelligence}, vol.~38, no.~3, pp. 417--430, 2016.

\bibitem{xu2017unified}
C.~Xu, Z.~Lin, and H.~Zha, ``A unified convex surrogate for the {Schatten-p}
  norm.'' in \emph{AAAI Conference on Artificial Intelligence}, 2017, pp.
  926--932.

\bibitem{Wang2013Provable}
Y.-X. Wang, H.~Xu, and C.~Leng, ``Provable subspace clustering: When
  \protect{LRR} meets \protect{SSC},'' in \emph{Advances in Neural Information
  Processing Systems}, 2013, pp. 64--72.

\bibitem{Lai2014Efficient}
H.~Lai, Y.~Pan, C.~Lu, Y.~Tang, and S.~Yan, ``Efficient {k-Support} matrix
  pursuit,'' in \emph{European Conference on Computer Vision}, 2014, pp.
  617--631.

\bibitem{Kim2016Robust}
E.~Kim, M.~Lee, and S.~Oh, ``Robust {Elastic-Net} subspace representation,''
  \emph{IEEE Transactions on Image Processing}, vol.~25, no.~9, pp. 4245--4259,
  2016.

\bibitem{Lu2015Correlation}
C.~Lu, J.~Feng, Z.~Lin, and S.~Yan, ``Correlation adaptive subspace
  segmentation by {Trace Lasso},'' in \emph{IEEE Conference on Computer Vision
  and Pattern Recognition}, 2015, pp. 1345--1352.

\bibitem{Xu2015Reweighted}
J.~Xu, K.~Xu, K.~Chen, and J.~Ruan, ``Reweighted sparse subspace clustering,''
  \emph{Computer Vision and Image Understanding}, vol. 138, pp. 25--37, 2015.

\bibitem{abin2018querying}
A.~A. Abin, ``Querying beneficial constraints before clustering using facility
  location analysis,'' \emph{IEEE Transactions on Cybernetics}, vol.~48, no.~1,
  pp. 312--323, 2018.

\bibitem{Guo2013Spatial}
Y.~Guo, J.~Gao, and F.~Li, ``Spatial subspace clustering for hyperspectral data
  segmentation,'' in \emph{International Conference on Digital Information
  Processing and Communications}, 2013, pp. 180--190.

\bibitem{wang2017constrained}
J.~Wang, X.~Wang, F.~Tian, C.~H. Liu, and H.~Yu, ``Constrained low-rank
  representation for robust subspace clustering,'' \emph{IEEE Transactions on
  Cybernetics}, vol.~47, no.~12, pp. 4534--4546, 2017.

\bibitem{yang2015multitask}
Y.~Yang, Z.~Ma, Y.~Yang, F.~Nie, and H.~T. Shen, ``Multitask spectral
  clustering by exploring intertask correlation,'' \emph{IEEE Transactions on
  Cybernetics}, vol.~45, no.~5, pp. 1083--1094, 2015.

\bibitem{Liu2013Robust}
G.~Liu, Z.~Lin, S.~Yan, J.~Sun, Y.~Yu, and Y.~Ma, ``Robust recovery of subspace
  structures by low-rank representation,'' \emph{IEEE Transactions on Pattern
  Analysis and Machine Intelligence}, vol.~35, no.~1, pp. 171--184, 2013.

\bibitem{Favaro2011A}
P.~Favaro, R.~Vidal, and A.~Ravichandran, ``A closed form solution to robust
  subspace estimation and clustering,'' in \emph{IEEE Conference on Computer
  Vision and Pattern Recognition}, 2011, pp. 1801--1807.

\bibitem{Li2014SCAMS}
Z.~Li, L.~F. Cheong, and S.~Z. Zhou, ``{SCAMS}: Simultaneous clustering and
  model selection,'' in \emph{IEEE Conference on Computer Vision and Pattern
  Recognition}, 2014, pp. 264--271.

\bibitem{Ester1996A}
M.~Ester, H.~P. Kriegel, J.~Sander, and X.~Xu, ``A density-based algorithm for
  discovering clusters in large spatial databases with noise,'' in
  \emph{International Conference on Knowledge Discovery and Data Mining}, 1996,
  p. 226–231.

\bibitem{Beier2015Fusion}
T.~Beier, F.~A. Hamprecht, and J.~H. Kappes, ``Fusion moves for correlation
  clustering,'' in \emph{IEEE Conference on Computer Vision and Pattern
  Recognition}, 2015, pp. 3507--3516.

\bibitem{lutensor}
C.~Lu, J.~Feng, Y.~Chen, W.~Liu, Z.~Lin, and S.~Yan, ``Tensor robust principal
  component analysis: Exact recovery of corrupted low-rank tensors via convex
  optimization,'' in \emph{IEEE Conference on Computer Vision and Pattern
  Recognition}, 2016, pp. 2080--2088.

\bibitem{Purkait2014Clustering}
P.~Purkait, T.~J. Chin, H.~Ackermann, and D.~Suter, ``Clustering with
  hypergraphs: The case for large hyperedges,'' \emph{IEEE Transactions on
  Pattern Analysis and Machine Intelligence}, vol.~39, no.~9, pp. 1697--1711,
  2017.

\bibitem{li2017graph}
X.~Li, G.~Cui, and Y.~Dong, ``Graph regularized non-negative low-rank matrix
  factorization for image clustering,'' \emph{IEEE Transactions on
  Cybernetics}, vol.~47, no.~11, pp. 3840--3853, 2017.

\bibitem{Sch2006Learning}
B.~Schölkopf, J.~Platt, and T.~Hofmann, ``Learning with hypergraphs:
  Clustering, classification, and embedding,'' in \emph{Advances in Neural
  Information Processing Systems}, 2006, pp. 1601--1608.

\bibitem{lee2015membership}
M.~Lee, J.~Lee, H.~Lee, and N.~Kwak, ``Membership representation for detecting
  block-diagonal structure in low-rank or sparse subspace clustering.'' in
  \emph{IEEE Conference on Computer Vision and Pattern Recognition}, 2015, pp.
  1648--1656.

\bibitem{Belkin2001Laplacian}
M.~Belkin and P.~Niyogi, ``Laplacian eigenmaps and spectral techniques for
  embedding and clustering,'' in \emph{Advances in Neural Information
  Processing Systems}, 2002, pp. 585--591.

\bibitem{NIPS_Manifold}
E.~Elhamifar and R.~Vidal, ``Sparse manifold clustering and embedding,'' in
  \emph{Advances in Neural Information Processing Systems}, 2011, pp. 55--63.

\bibitem{Li2009Learning}
C.~Li, J.~Guo, and H.~Zhang, ``Learning bundle manifold by double neighborhood
  graphs,'' in \emph{Asian Conference on Computer Vision}, 2009, pp. 321--330.

\bibitem{von2007tutorial}
U.~Von~Luxburg, ``A tutorial on spectral clustering,'' \emph{Statistics and
  Computing}, vol.~17, no.~4, pp. 395--416, 2007.

\bibitem{Georghiades2001From}
A.~S. Georghiades, P.~N. Belhumeur, and D.~J. Kriegman, ``From few to many:
  Illumination cone models for face recognition under variable lighting and
  pose,'' \emph{IEEE Transactions on Pattern Analysis and Machine
  Intelligence}, vol.~23, no.~6, pp. 643--660, 2001.

\bibitem{Rate2011Columbia}
S.~A. Nene, S.~K. Nayar, and H.~Murase, ``Columbia object image library
  ({COIL}-20),'' \emph{Columbia Universty, Tech. Rep. CUCS-005-96}, 1996.

\bibitem{tron2007benchmark}
R.~Tron and R.~Vidal, ``A benchmark for the comparison of {3-D} motion
  segmentation algorithms,'' in \emph{IEEE Conference on Computer Vision and
  Pattern Recognition}, 2007, pp. 1--8.

\bibitem{Li2003Clustering}
M.~Li, X.~Chen, X.~Li, and B.~Ma, ``Clustering by compression,'' in \emph{IEEE
  International Symposium on Information Theory}, vol.~51, no.~4, 2003, pp.
  1523--1545.

\end{thebibliography}
	
\end{document}